\pdfoutput=1

\documentclass[11pt]{article}

\usepackage{arabtex}

\usepackage{EMNLP2022}

\usepackage{times}
\usepackage{latexsym}

\usepackage{booktabs}
\usepackage{graphics}
\usepackage{graphicx}
\usepackage{adjustbox}
\usepackage[normalem]{ulem}
\useunder{\uline}{\ul}{}
\usepackage{longtable}
\usepackage{CJKutf8}

\usepackage[T2A, T1]{fontenc}

\usepackage[utf8]{inputenc}
\usepackage{utf8}

\usepackage{microtype}

\usepackage{inconsolata}

\usepackage{fontawesome5}

\usepackage{makecell}

\usepackage{blindtext}

\usepackage{multirow}

\usepackage{diagbox}

\title{Revisiting non-English Text Simplification: \\ A Unified Multilingual Benchmark}

\author{Michael J. Ryan, Tarek Naous, Wei Xu \\
School of Interactive Computing \\
  Georgia Institute of Technology \\
  {\small \texttt{\{michaeljryan, tareknaous\}@gatech.edu; wei.xu@cc.gatech.edu}} \\}

\begin{document}
\setcode{utf8}
\maketitle

\begin{abstract}

Recent advancements in high-quality, large-scale English resources have pushed the frontier of English Automatic Text Simplification (ATS) research. However, less work has been done on multilingual text simplification due to the lack of a diverse evaluation benchmark that covers complex-simple sentence pairs in many languages. This paper introduces the {\sc MultiSim} benchmark, a collection of 27 resources in 12 distinct languages containing over 1.7 million complex-simple sentence pairs. This benchmark will encourage research in developing more effective multilingual text simplification models and evaluation metrics. Our experiments using {\sc MultiSim} with pre-trained multilingual language models reveal exciting performance improvements from multilingual training in non-English settings. We observe strong performance from Russian in zero-shot cross-lingual transfer to low-resource languages. We further show that few-shot prompting with BLOOM-176b achieves comparable quality to reference simplifications outperforming fine-tuned models in most languages. We validate these findings through human evaluation.\footnote{Code and Data available at \url{https://github.com/XenonMolecule/MultiSim}}

\end{abstract}

\section{Introduction}

Automatic text simplification (ATS) is the task of reducing the complexity of a text without changing its original content and meaning \citep{10.1145/3442695}. ATS has many applications, from making a text easier to read for people with reading and cognitive disabilities \citep{stajner-2021-automatic} and second language learners \citep{petersen2007text} to reducing the complexity of medical texts for easier understanding by the general public \citep{10.1145/3308558.3313630}.  For better accessibility to diverse communities, this technology should be available without language barriers. 

\begin{figure}[t!]
    \centering
    \includegraphics[width=0.99\linewidth]{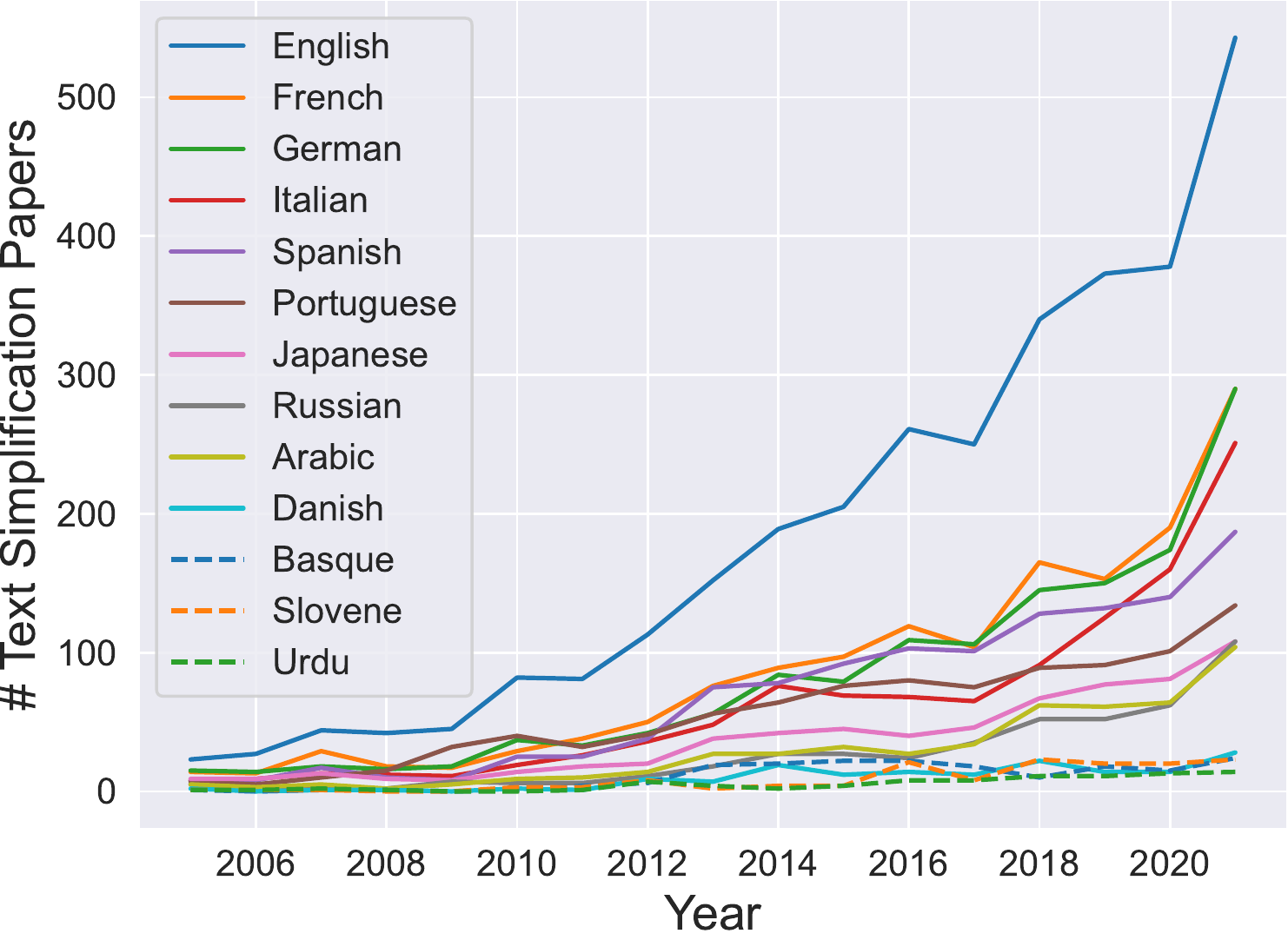}
    \caption{Papers published each year with content related to text simplification and a specific language according to Google Scholar.  The quantity of English text simplification work vastly exceeds all other languages.}
    \label{fig:language_over_time}
\end{figure}

Much of the recent success in English text simplification comes from large parallel corpora of texts with the same content written using both complicated and simple sentences \citep{xu-etal-2015-problems, jiang-etal-2020-neural, alva-manchego-etal-2020-asset}. These resources enable the training of large language models for ATS in English \citep{scarton-specia-2018-learning, martin-etal-2020-controllable, omelianchuk-etal-2021-text}. ATS research in other languages has received much less attention  \citep{martin-etal-2022-muss}. Figure \ref{fig:language_over_time} shows that the growth of English text simplification research outpaces progress in other languages.

A diverse multilingual benchmark is essential for a more comprehensive evaluation of multilingual simplification methods, pre-trained models, and evaluation metrics. The lack of a multilingual benchmark that covers a set of high, medium, and low-resource languages belonging to different scripts and language families hinders advancement in multilingual ATS. In this paper, we address this gap in the field by introducing the {\sc MultiSim} benchmark that covers 27 text simplification datasets (complex-simple pairs) in 12 different languages. {\sc MultiSim} consists of a collection of datasets from the literature that we unify into a single format for easier accessibility to the research community. In summary, our main contributions are as follows: 

\begin{enumerate}
\item We present a comprehensive literature survey of all existing multilingual text simplification corpora, created via several methodologies categorized into four main approaches (\S \ref{sec:parallel-simplification-corpora}).

\item We release the {\sc MultiSim} benchmark for multilingual text simplification, containing 1,749,056 simple-complex sentence pairs in 12 different languages.  To our knowledge, this is the first multilingual benchmark for text simplification. (\S \ref{sec:MultiSim-Benchmark}).

\item We run various experiments using pre-trained multilingual language models and analyze their effectiveness in few-shot learning and cross-lingual transfer for challenging cases of low-resource languages or domain-specific simplification (\S \ref{sec:Experiments}). Our results highlight the benefits of domain and language script match for zero-shot transfer. We find that few-shot prompting large language models produces high-quality simplifications in both high and low-resource languages (\S \ref{sec:results}). We validate these findings with human evaluation (\S \ref{sec:manual-analysis}).

\end{enumerate}



\section{Related Works}

\subsection{Multilingual Benchmarks}

Recently researchers have released several multilingual benchmarks to assess that models work well not only in the high-resource settings where they are trained but in all languages.  \textit{XTREME-R} \citep{hu2020xtreme} is a multitask benchmark across 50 languages.  The benchmark focuses on classification, question answering, structured prediction, and retrieval. Another text classification benchmark is \textit{XGLUE} \citep{liang-etal-2020-xglue}, which covers 11 diverse tasks in 19 languages.  Finally, the new \textit{XTREME-UP} benchmark \citep{ruder2023xtreme} evaluates 88 under-represented languages on 9 tasks from machine translation to OCR to autocomplete.

Single task multilingual benchmarks exist for NLI \citep{conneau-etal-2018-xnli}, QA \citep{lewis-etal-2020-mlqa, 10.1162/tacl_a_00433}, causal reasoning \citep{ponti2020xcopa}, semantic similarity  \citep{10.1162/coli_a_00391}, style transfer \citep{briakou-etal-2021-ola}, fact checking \citep{gupta-srikumar-2021-x}, fairness \citep{chalkidis-etal-2022-fairlex}, stance classification \citep{zheng2022stanceosaurus}, text summarization \citep{giannakopoulos-etal-2015-multiling, ladhak-etal-2020-wikilingua, scialom-etal-2020-mlsum}, readability \citep{naous2023massively} and more \citep{gretter-2014-euronews, MEILICKE201262, li2020mtop, raganato2020xl}.  To date, no such benchmarks exist for multilingual text simplification.

\subsection{Multilingual Text Simplification}

The most common approach to automatic text simplification is training a statistical or neural sequence-to-sequence generation model on parallel simplification corpora (\S \ref{sec:parallel-simplification-corpora}). Besides in English, researchers have done this in Brazilian Portuguese \citep{Specia2010TranslatingFC}, German \citep{sauberli-etal-2020-benchmarking, battisti-etal-2020-corpus}, Spanish \citep{stajner-etal-2015-automatic, tajner2014TranslatingSF}, French \citep{cardon-grabar-2020-french}, Japanese \citep{goto-etal-2015-japanese, 8300618}, Danish \citep{klerke-sogaard-2013-simple}, and Russian \citep{shatilov2021sentence, Fenogenova2021TextSW}.

Zero-shot and unsupervised learning are promising directions for multilingual text simplification. \citet{mallinson-etal-2020-zero} showed zero-shot simplification in German using a German decoder on a transformer architecture trained on English simplification. Additionally, \citet{martin-etal-2022-muss} showed that mining a massive amount of paraphrases and simplifications as training data was sufficient to achieve state-of-the-art performance in English, Spanish, and French text simplification.

\begin{table*}[h!]
\small
\setlength{\tabcolsep}{2pt}
\renewcommand{\arraystretch}{0.5}
\centering
\resizebox{1.0\textwidth}{!}{%
\begin{tabular}{l|cccccccc}
\toprule
\thead{\textbf{Corpus}} & \thead{\textbf{Source(s)}} & \textbf{\thead{Simplification \\ Author}} & 
\textbf{\thead{Collection \\ Strategy}} & 
\textbf{\thead{Alignment\\Level}} &
\textbf{\thead{Sentence \\ Aligned}} & \textbf{\thead{Complex \\ Sentences}} & \textbf{\thead{Simple \\ Sentences}} & \thead{\textbf{Access}} \\
\midrule
\textbf{Arabic Corpora}  & \multicolumn{7}{c}{} \\
\textit{Saaq al-Bambuu} \citep{khallaf2022towards} & \faIcon{book} & writer & \faIcon{star} & sentence & auto & 2,980 & 2,980 & private \\
\midrule
\textbf{Basque Corpora} & \multicolumn{7}{c}{} \\
\textit{CBST} \citep{10.1007/s10579-017-9407-6} & \faIcon{flask} & translator, teacher & \faIcon{pen} & document & manual & 458 & 591 & on request \\
\midrule
\textbf{Brazilian Portuguese Corpora} & \multicolumn{7}{c}{} \\
\textit{PorSimples} \citep{aluisio-gasperin-2010-fostering} & \faIcon[regular]{newspaper} \faIcon{flask} & linguist & \faIcon{pen} & document & manual & 7,902 & 10,174 & on request \\
\midrule
\textbf{Danish Corpora} & \multicolumn{7}{c}{} \\
\textit{DSim} \citep{klerke-sogaard-2012-dsim} & \faIcon[regular]{newspaper} & journalists & \faIcon{star} & sentence & auto & 47,887 & 60,528 & on request \\
\midrule
\textbf{English Corpora$\dagger$} & \multicolumn{7}{c}{} \\
\textit{ASSET} \citep{alva-manchego-etal-2020-asset} & \faIcon{wikipedia-w} & crowdsource & \faIcon{pen} & sentence & manual & 2,359 & 23,590 & open source \\
\textit{Newsela EN} \citep{xu-etal-2015-problems} & \faIcon[regular]{newspaper} & experts & \faIcon{star} & document & auto & 393,798 & 402,222 & on request \\
\textit{Wiki-Auto} \citep{jiang-etal-2020-neural} & \faIcon{wikipedia-w} & crowdsource & \faIcon{cog} & document & auto & 10,144,476 & 1,241,671 & open source \\
\midrule
\textbf{French Corpora} & \multicolumn{7}{c}{} \\
\textit{Alector} \citep{gala-etal-2020-alector} & \faIcon{globe} \faIcon{flask} & experts & \faIcon{pen} & document & NA & 1,230 & 1,192 & open source \\
\textit{CLEAR} \citep{grabar-cardon-2018-clear} & \faIcon{wikipedia-w} \faIcon{file-medical} & crowdsource, experts & \faIcon{cog} & sentence & auto & 4,596 & 4,596 & open source \\
\textit{WikiLarge FR} \citep{cardon-grabar-2020-french} & \faIcon{wikipedia-w} & crowdsource & \faIcon{language} & sentence & auto & 307,067 & 308,409 & open source \\
\midrule
\textbf{German Corpora} & \multicolumn{7}{c}{} \\
\textit{GEOLinoTest} \citep{mallinson-etal-2020-zero} & \faIcon[regular]{newspaper} & linguist & \faIcon{pen} & sentence & manual & 1,198 & 1,198 & open source \\
\textit{German News} \citep{sauberli-etal-2020-benchmarking} & \faIcon[regular]{newspaper} & news agency & \faIcon{star} & document & auto & 15,239 & 14,344 & on request \\
\textit{Klexikon} \citep{aumiller-gertz-2022-klexikon} & \faIcon{wikipedia-w} & crowdsource & \faIcon{cog} & document & NA & 771,059 & 96,870 & open source \\
\textit{Simple Patho} \citep{trienespatient} & \faIcon{file-medical} & medical students & \faIcon{pen} & paragraph & manual & 22,191 & 26,551 & private \\
\textit{Simple German} \citep{battisti-etal-2020-corpus} & \faIcon{globe} & government & \faIcon{star} & document & auto & 12,806 & 8,400 & on request* \\
\textit{TextComplexityDE} \citep{https://doi.org/10.48550/arxiv.1904.07733} & \faIcon{wikipedia-w} & native speaker & \faIcon{pen} & document & manual & 250 & 250 & open source \\
\midrule
\textbf{Italian Corpora} & \multicolumn{7}{c}{} \\
\textit{AdminIT} \citep{miliani-etal-2022-neural} & \faIcon{gavel} & researchers & \faIcon{pen} & sentence & manual & 777 & 763 & open source \\
\textit{SIMPITIKI Wiki} \citep{Tonelli2016} & \faIcon{wikipedia-w} & crowdsource & \faIcon{cog} & sentence & manual & 575 & 575 & open source \\
\textit{PaCCSS-IT} \citep{brunato-etal-2016-paccss} & \faIcon{globe} & crowdsource & \faIcon{cog} & sentence & auto & 63,006 & 63,006 & open source \\
\textit{Teacher} \citep{brunato-etal-2015-design} & \faIcon{globe} & \makecell[c]{teachers} & \faIcon{pen} & document & manual & 204 & 195 & open source \\
\textit{Terence} \citep{brunato-etal-2015-design} & \faIcon{book} & \makecell[c]{experts} & \faIcon{pen} & document & manual & 1,035 & 1,060 & open source \\
\midrule
\textbf{Japanese Corpora} & \multicolumn{7}{c}{} \\
\textit{EasyJapanese} \citep{maruyama-yamamoto-2018-simplified} & \faIcon[regular]{newspaper} \faIcon{globe} & students & \faIcon{pen} & sentence & manual & 50,000 & 50,000 & open source \\
\textit{EasyJapaneseExtended} \citep{katsuta-yamamoto-2018-crowdsourced} & \faIcon[regular]{newspaper} \faIcon{globe} & crowdsource & \faIcon{pen} & sentence & manual & 34,400 & 35,000 & open source \\
\textit{Japanese News} \citep{goto-etal-2015-japanese} & \faIcon[regular]{newspaper} & journalists, teachers & \faIcon{star} & document & auto & 13,356 & 13,356 & private \\
\midrule
\textbf{Russian Corpora} & \multicolumn{7}{c}{} \\
\textit{RuAdapt Encyclopedia} \citep{Dmitrieva2021Quantitative} & \faIcon{info-circle} & researchers & \faIcon{pen} & document & auto & 9,729 & 10,230 & open source \\
\textit{RuAdapt Fairytale} \citep{Dmitrieva2021Quantitative} & \faIcon{book} & researchers & \faIcon{pen} & document & auto & 310 & 404 & open source \\
\textit{RuAdapt Lit} \citep{dmitrieva-tiedemann-2021-creating} & \faIcon{book} & writers & \faIcon{pen} & document & auto & 24,152 & 28,259 & on request \\
\textit{RSSE} \citep{sakhovskiy2021rusimplesenteval} & \faIcon{wikipedia-w} & crowdsource & \faIcon{pen} & sentence & manual & 2,000 & 6,804 & open source \\
\textit{RuWikiLarge} \citep{sakhovskiy2021rusimplesenteval} & \faIcon{wikipedia-w} & crowdsource & \faIcon{language} & sentence & auto & 278,499 & 289,788 & on request \\
\midrule
\textbf{Slovene Corpora} & \multicolumn{7}{c}{} \\
\textit{SloTS} \citep{11356/1682} & \faIcon{book} & experts & \faIcon{star} & sentence & manual & 1,181 & 1,287 & open source \\
\midrule
\textbf{Spanish Corpora} & \multicolumn{7}{c}{} \\
\textit{FIRST} \cite{orasan2013text} & \faIcon{book} \faIcon[regular]{newspaper} \faIcon{file-medical} &  experts & \faIcon{pen} & document & manual & 320 & 332 & private \\
\textit{Newsela ES} \citep{xu-etal-2015-problems} & \faIcon[regular]{newspaper} & experts & \faIcon{star} & document & auto & 46,256 & 45,519 & on request \\
\textit{Simplext} \citep{10.1145/2738046} & \faIcon[regular]{newspaper} & researchers & \faIcon{pen} & document & manual & 1,108 & 1,742 & on request \\
\midrule
\textbf{Urdu Corpora} & \multicolumn{7}{c}{} \\
\textit{SimplifyUREval} \citep{qasmi-etal-2020-simplifyur} & \faIcon{book} \faIcon[regular]{newspaper} & expert & \faIcon{pen} & sentence& manual & 500 & 736 & open source \\
\bottomrule
\end{tabular}}
\caption{\label{table:corpora-stats}
Important properties of text simplification parallel corpora. $\dagger$Common English corpora included for comparison. Many other English corpora omitted. *Only scripts to replicate the corpus are available upon request.  Simple German results differ from original paper because of changes to availability of online articles.  \textit{Sources}: \faIcon{book} Literature, \faIcon{flask} Science Communications, \faIcon[regular]{newspaper} News, \faIcon{wikipedia-w}ikipedia, \faIcon{globe} Websites, \faIcon{file-medical} Medical Documents, \faIcon{gavel} Government, \faIcon{info-circle} Encyclopedic.  \textit{Collection Strategies}: \faIcon{cog} Automatic, \faIcon{language} Translation, \faIcon{pen} Annotator, \faIcon{star} Target Audience Resource.
}
\end{table*}

\begin{figure}[t!]
    \centering
    \includegraphics[width=1.0\linewidth]{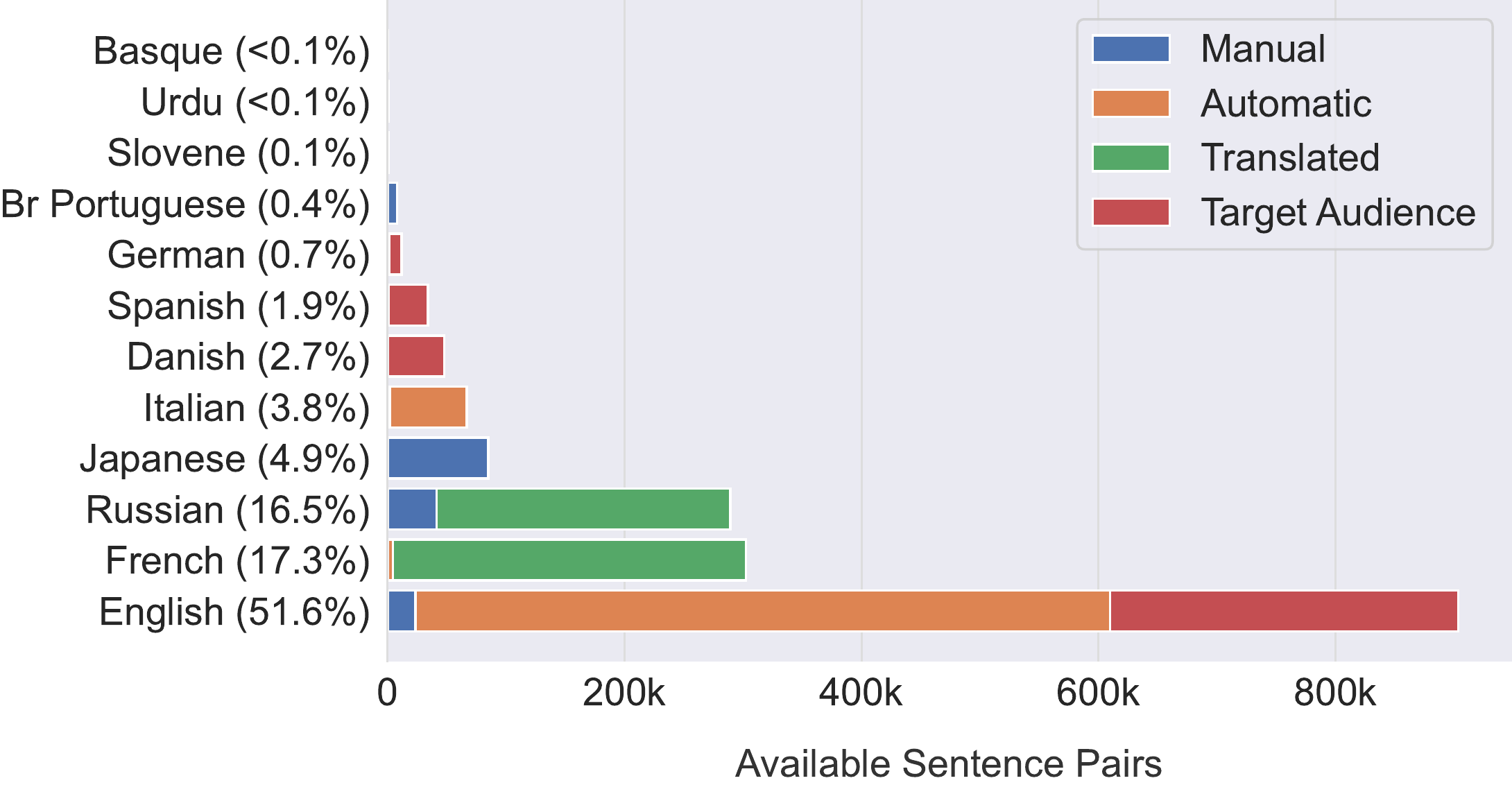}
    \caption{Data availability for text simplification in all languages partitioned on collection strategy.  Despite only including three of the most common English datasets, English resources outnumber all other language resources combined.}
    \label{fig:datasize}
\end{figure}

\section{Parallel Simplification Corpora}
\label{sec:parallel-simplification-corpora}
So far, 31 parallel simplification corpora exist in non-English languages. We organize the discussion of these corpora by their creation strategy. Figure \ref{fig:datasize} shows the amount of data in each language divided by collection strategy. Table \ref{table:corpora-stats} summarizes the details of each corpus stratified by language. We provide more detail on each corpus in Appendix \ref{sec:summary}.

\subsection{Manual Simplification}

Manual simplification is the most widely-used method for crafting a monolingual parallel simplification corpus. For manual simplification, annotators ranging from experts to crowdsourced workers to the researchers themselves use simplification guidelines \citep{10.1145/1410140.1410191, Siddharthan2004, 10.1007/s10579-017-9407-6} to simplify complex documents manually. Researchers have used this methodology to create 17 resources in 10 different languages. Manual simplification gives researchers control over the types of simplification operations in the dataset. However, guiding annotators using rules can result in unnatural simplifications. \citet{Stajner2014Translating} found that relaxing guidelines led to more effective simplifications.

\subsection{Automatic Collection}

Manual simplification at scale is a costly and time-consuming process. An alternative approach is to use automatic collection methods, which leverage various online knowledge bases or web sources to increase the size of these resources. However, this can come at the cost of sacrificing quality, as automatically paired sentences may not always be an exact complex-simple match. Additionally, it can be challenging to control the level of simplification.

One common source for automatic collection is Wikipedia, which is available in many languages and often has both original and simplified versions with the same content.\footnote{\url{https://simple.wikipedia.org/wiki/Main_Page}} This has been used to build parallel simplification corpora \citep{jiang-etal-2020-neural, Tonelli2016, cardon-grabar-2019-parallel}, where researchers match articles on the same topic from the regular and simple Wikipedia versions.  They then leverage automatic aligners such as a neural CRF aligner \citep{jiang-etal-2020-neural} or CATS \citep{stajner-etal-2018-cats} to find matching sentences between the two articles.

Other sources of automatically simplified sentences include web scrapes like Common Crawl \citep{wenzek-etal-2020-ccnet}. For web scrapes, sentence embeddings \citep{heffernan2022bitext} are used to find similar sentences to pair up. Then additional filtering is applied to ensure the sentences are not exact matches. A readability measure can be used to ensure that one sentence is simpler than the other. This was the strategy used to create PaCCSS-IT \citep{brunato-etal-2016-paccss} in Italian and MUSS \citep{martin-etal-2022-muss} in English, Spanish, and French.

\subsection{Machine Translation}
Some datasets are machine translations of existing large resources. For example, the French WikiLargeFR \citep{cardon-grabar-2020-french} and Russian RuWikiLarge \citep{sakhovskiy2021rusimplesenteval} are both machine translations of the English WikiLarge corpus \citep{zhang-lapata-2017-sentence}.  While this allows for large resources in multiple languages, it has two significant drawbacks. Firstly, the final dataset lacks the cultural identity of naturally occurring data in the target language. Secondly, machine translation errors can be introduced in the process, potentially impacting the dataset's quality.

\subsection{Target Audience Resources}

The final monolingual parallel corpora category is resources created for specific target audiences, such as individuals with lower literacy levels or second language learners. These resources typically have the highest quality, but they can be expensive to produce and therefore are relatively rare. There are currently target audience resources available in seven languages.

One company that specializes in creating high-quality target audience resources in English and Spanish is Newsela\footnote{\url{https://newsela.com}} \citep{xu-etal-2015-problems}. Founded in 2013, Newsela is a Series D startup that has attracted over \$100 million in funding to support its goal of promoting meaningful classroom learning at all levels. The company employs a team of content producers with extensive teaching experience in K-12 education to train and manage a network of freelance writers who create the simplifications.

Several national news agencies have established systems for creating simplified versions of their articles. For instance, News Web Easy\footnote{\url{https://www3.nhk.or.jp/news/easy/}} in Japan is a division of the Japan Broadcasting Corporation, a public media organization. News Web Easy targets Japanese second language learners and primary and secondary school students \citep{goto-etal-2015-japanese}. The German government funds the Austrian Press Agency to produce TopEasy\footnote{\url{https://science.apa.at/nachrichten-leicht-verstandlich/}} \citep{sauberli-etal-2020-benchmarking}, a simplified version of their news published each weekday. Similarly, the publicly funded Danish Broadcasting Corporation (DR) offers simplified versions of their stories called DR Ligetil\footnote{\url{https://www.dr.dk/ligetil}} (Straightforward) \citep{klerke-sogaard-2012-dsim}.


\label{sec:datasets}

\section{The MultiSim Benchmark}
\label{sec:MultiSim-Benchmark}

We release the {\sc MultiSim} Benchmark, a collection of 27 parallel simplification corpora in 12 languages and 4 Scripts. 18 of these corpora are open-sourced and are available online. For 9 corpora, permission must be obtained from the original authors. We provide data loaders for these resources.

\subsection{Languages}

We included all languages with open-source parallel sentence-aligned text simplification corpora. This covers eight languages: English (en), French (fr), German (de), Italian (it), Japanese (ja), Russian (ru), Slovene (sl), and Urdu (ur). Six languages have corpora available on request. We provide data loaders and splits to make these resources compatible with the {\sc MultiSim} benchmark. Such resources exist in Basque (eu), Brazilian Portuguese (pt-br), Danish (da), German (de), Russian (ru), and Spanish (es). Some resources are entirely private due to copyright protection or data-sharing permissions. These resources are in Arabic (ar), German (de), Japanese (ja), and Spanish (es).

\subsection{Domains}

The {\sc MultiSim} benchmark spans 8 domains. \textit{Literature} sources are simplified versions of novels. \textit{Science Communications} are popular science articles already written for public consumption and then rewritten at a simpler level. \textit{News} sources are simplified versions of articles and news stories. \textit{Wikipedia} sources are pulled from original and simple Wikipedia sites. \textit{Website} sources generally come from web scrapes like Common Crawl \citep{wenzek-etal-2020-ccnet} or specific target websites with original and simplified texts. \textit{Medical} documents are drug leaflets, clinical notes, and similar texts written for doctors but simplified so that an average patient could understand. \textit{Government} documents are taken from government policies and simplified to use more common vernacular. \textit{Encyclopedic} documents are informational texts like Wikipedia but from other encyclopedic sources.

\begin{table}
\small
\setlength{\tabcolsep}{2pt}
\renewcommand{\arraystretch}{0.5}
\centering
\resizebox{0.41\textwidth}{!}{%
\begin{tabular}{@{}llrrr@{}}
\toprule
Language & Dataset & \multicolumn{1}{c}{\#train} & \multicolumn{1}{c}{\#test} & \multicolumn{1}{c}{\#dev} \\ \midrule
\multicolumn{5}{c}{\textbf{Open Source}} \\ \midrule \midrule
\multirow{2}{*}{English} & WikiAuto & 576,126 & 5,012 & 5,012 \\
 & ASSET* & 20,000 & 3,590 & 0 \\ \midrule
\multirow{2}{*}{French} & WikiLargeFR* & 296,402 & 359 & 992 \\
 & CLEAR* & 4,196 & 100 & 300 \\ \midrule
\multirow{2}{*}{German} & GEOLino & 958 & 122 & 118 \\
 & TextCompDE & 200 & 25 & 25 \\ \midrule
\multirow{5}{*}{Italian} & PaCCSS-IT & 60,485 & 1,267 & 1,254 \\
 & Terence & 809 & 101 & 102 \\
 & AdminIT & 588 & 73 & 75 \\
 & Simpitiki & 460 & 56 & 59 \\
 & Teacher & 136 & 17 & 17 \\ \midrule
\multirow{2}{*}{Japanese} & Easy JA & 48,000 & 1,000 & 1,000 \\
 & Easy JA Ext* & 34,269 & 731 & 0 \\ \midrule
\multirow{3}{*}{Russian} & RuAdapt Ency & 7,782 & 982 & 965 \\
 & RSSE Corpus* & 3,406 & 3,398 & 0 \\
 & RuAdapt Fairy & 248 & 31 & 31 \\ \midrule
 Slovene & SloTS* & 749 & 96 & 94 \\ \midrule
Urdu & SimplifyUR & 594 & 68 & 74 \\ \midrule
 & Total & 1,055,408 & 17,028 & 10,118 \\ \midrule 
\multicolumn{5}{c}{\textbf{Dataloaders Available (Data on Request)}} \\ \midrule \midrule
Basque & CBST & 361 & 46 & 46 \\ \midrule
Br Portuguese & PorSimples & 6,290 & 790 & 784 \\ \midrule
Danish & DSim Corpus & 45,885 & 997 & 1,005 \\ \midrule
English & Newsela EN & 291,969 & 991 & 1,008 \\ \midrule
German & German News & 8,186 & 1,024 & 1,023 \\ \midrule
\multirow{2}{*}{Russian} & RuWikiLarge* & 246,978 & 365 & 768 \\
 & RuAdapt Lit & 22,152 & 1,000 & 1,000 \\ \midrule
\multirow{2}{*}{Spanish} & Newsela ES & 30,910 & 1,001 & 1,001 \\
 & Simplext & 737 & 92 & 93 \\ \midrule
 & Total & 653,468 & 6,306 & 6,728 \\ \bottomrule
\end{tabular}
}
\caption{\label{table:multisim-splits}
{\sc MultiSim} splits. *Original splits preserved
}
\end{table}

\subsection{Pre-processing and Splitting}

For any resource that provided a train, test, dev split, we include the original split of the data in our collection. Otherwise, we randomly divided all sentence pairs into train, test, and dev sets. For resources under 10,000 sentence pairs, we used 80\%/10\%/10\% splits. For resources above 10,000 sentence pairs, we randomly sampled about 1,000 sentences each for the test/dev sets. For resources above 500,000 sentence pairs (WikiAuto), we randomly sampled about 5,000 sentences each for the test/dev sets. We report split sizes in Table \ref{table:multisim-splits}.

Since several resources in the benchmark come from overlapping domains (i.e., Wikipedia, Web, News), repeat sentences exist between the original datasets. To fix this, we identified overlapping sentences and ensured they fell in the same split by swapping with randomly sampled sentence pairs. We repeated this process until all splits were completely independent.

\begin{table*}
\small
\setlength{\tabcolsep}{2pt}
\renewcommand{\arraystretch}{0.5}
\centering
\resizebox{0.98\textwidth}{!}{%
\begin{tabular}{@{}lrr|ccccc|ccccc@{}}
\toprule
\multicolumn{3}{l|}{} & \multicolumn{5}{c|}{\textbf{BLEU}} & \multicolumn{5}{c}{\textbf{SARI}} \\ \midrule
 & \multicolumn{1}{l}{} & \multicolumn{1}{l|}{} & \multicolumn{2}{c|}{\texttt{Baseline}} & \multicolumn{3}{c|}{\texttt{Finetune}} & \multicolumn{2}{c|}{\texttt{Baseline}} & \multicolumn{3}{c}{\texttt{Finetune}} \\ \midrule
\texttt{Lang} & \multicolumn{1}{r|}{\texttt{Dataset}} & \texttt{Size} & \texttt{Identity} & \multicolumn{1}{c|}{\texttt{Trunc}} & \texttt{Single} & \texttt{Lang} & \texttt{All} & \texttt{Identity} & \multicolumn{1}{c|}{\texttt{Trunc}} & \texttt{Single} & \texttt{Lang} & \texttt{All} \\ \midrule
\texttt{eu} & \multicolumn{1}{r|}{\texttt{CBST}} & \texttt{218} & \;\;72.02\;\; & \multicolumn{1}{c|}{\;\;57.87\;\;} & --- & --- & \;\;66.75\;\; & \;\;23.46\;\; & \multicolumn{1}{c|}{\;\;32.58\;\;} & --- & --- & \textbf{\;\;32.83\;\;} \\ \midrule
\texttt{ur} & \multicolumn{1}{r|}{\texttt{SimplifyUR}} & \texttt{470} & 58.85 & \multicolumn{1}{c|}{41.11} & --- & --- & 56.23 & 24.84 & \multicolumn{1}{c|}{31.30} & --- & --- & \textbf{51.74} \\ \midrule
\texttt{sl} & \multicolumn{1}{r|}{\texttt{SloTS}} & \texttt{188} & 7.76 & \multicolumn{1}{c|}{6.09} & --- & --- & 7.63 & 5.93 & \multicolumn{1}{c|}{19.03} & --- & --- & \textbf{30.52} \\ \midrule
\texttt{pt-br} & \multicolumn{1}{r|}{\texttt{PorSimples}} & \texttt{1,949} & 73.67 & \multicolumn{1}{c|}{51.93} & --- & --- & 63.85 & 28.21 & \multicolumn{1}{c|}{31.25} & --- & --- & \textbf{44.27} \\ \midrule
\multirow{3}{*}{\texttt{de}} & \multicolumn{1}{r|}{\texttt{TextCompDE}} & \texttt{144} & 26.77 & \multicolumn{1}{c|}{19.98} & --- & --- & 24.53 & 15.42 & \multicolumn{1}{c|}{26.81} & --- & --- & \textbf{41.15} \\
 & \multicolumn{1}{r|}{\texttt{GEOLino}} & \texttt{437} & 69.86 & \multicolumn{1}{c|}{50.03} & --- & --- & 71.90 & 27.45 & \multicolumn{1}{c|}{30.70} & --- & --- & \textbf{50.75} \\
 & \multicolumn{1}{r|}{\texttt{GermanNews}} & \texttt{1,748} & 7.29 & \multicolumn{1}{c|}{7.13} & --- & --- & 6.57 & 5.61 & \multicolumn{1}{c|}{17.69} & --- & --- & \textbf{31.58} \\ \midrule
\multirow{2}{*}{\texttt{es}} & \multicolumn{1}{r|}{\texttt{Simplext}} & \texttt{157} & 13.91 & \multicolumn{1}{c|}{13.15} & --- & \;\;14.42\;\; & 12.25 & 7.94 & \multicolumn{1}{c|}{20.27} & --- & \;\;19.91\;\; & \textbf{32.68} \\
 & \multicolumn{1}{r|}{\texttt{NewselaES}} & \texttt{17,022} & 58.18 & \multicolumn{1}{c|}{43.06} & \;\;51.78\;\; & 53.12 & 48.94 & 24.21 & \multicolumn{1}{c|}{31.64} & \;\;29.89\;\; & 28.56 & \textbf{35.36} \\ \midrule
\texttt{da} & \multicolumn{1}{r|}{\texttt{DSim}} & \texttt{25,524} & 31.39 & \multicolumn{1}{c|}{28.85} & 33.66 & 33.66 & 27.25 & 16.25 & \multicolumn{1}{c|}{26.10} & 31.40 & 31.40 & \textbf{38.44} \\ \midrule
\multirow{6.5}{*}{\texttt{it}} & \multicolumn{1}{r|}{\texttt{Simpitiki}} & \texttt{24} & 95.23 & \multicolumn{1}{c|}{74.48} & --- & 24.40 & 36.28 & 32.45 & \multicolumn{1}{c|}{32.00} & --- & 20.10 & \textbf{24.27} \\
& \multicolumn{1}{r|}{\texttt{Teacher}} & \texttt{83} & 34.49 & \multicolumn{1}{c|}{29.05} & --- & 32.21 & 29.76 & 17.41 & \multicolumn{1}{c|}{27.75} & --- & 29.98 & \textbf{30.97} \\
 & \multicolumn{1}{r|}{\texttt{AdminIT}} & \texttt{114} & 52.50 & \multicolumn{1}{c|}{45.63} & --- & 40.09 & 43.80 & 20.89 & \multicolumn{1}{c|}{28.22} & --- & 34.72 & \textbf{36.21} \\
 & \multicolumn{1}{r|}{\texttt{Terence}} & \texttt{394} & 67.24 & \multicolumn{1}{c|}{49.72} & --- & 59.33 & 50.65 & 26.83 & \multicolumn{1}{c|}{32.82} & --- & \textbf{37.77} & 36.92 \\
 & \multicolumn{1}{r|}{\texttt{PaCCSS-IT}} & \texttt{55,274} & 36.76 & \multicolumn{1}{c|}{28.77} & 49.57 & 48.31 & 42.87 & 18.14 & \multicolumn{1}{c|}{28.26} & \textbf{57.30} & 55.98 & 54.43 \\ \midrule
\multirow{2}{*}{\texttt{ja}} & \multicolumn{1}{r|}{\texttt{EasyJA}} & \texttt{27,600} & 58.09 & \multicolumn{1}{c|}{8.43} & 65.83 & 68.12 & 66.04 & 24.64 & \multicolumn{1}{c|}{24.28} & 67.36 & \textbf{70.95} & 70.11 \\
 & \multicolumn{1}{r|}{\texttt{EasyJAExt}} & \texttt{32,248} & 20.23 & \multicolumn{1}{c|}{0.00} & 33.07 & 35.67 & 31.50 & 9.00 & \multicolumn{1}{c|}{35.32} & 43.15 & 50.26 & \textbf{53.49} \\ \midrule
\multirow{6.5}{*}{\texttt{ru}} & \multicolumn{1}{r|}{\texttt{RuAdaptFairy}} & \texttt{97} & 12.56 & \multicolumn{1}{c|}{8.03} & --- & 13.11 & 11.01 & 10.63 & \multicolumn{1}{c|}{24.84} & --- & 23.77 & \textbf{26.55} \\
& \multicolumn{1}{r|}{\texttt{RuAdapt Ency}} & \texttt{1,450} & 84.15 & \multicolumn{1}{c|}{59.66} & --- & 76.06 & 61.83 & 29.90 & \multicolumn{1}{c|}{31.09} & --- & \textbf{34.73} & 34.40 \\
 & \multicolumn{1}{r|}{\texttt{RSSE}} & \texttt{1,477} & 38.23 & \multicolumn{1}{c|}{34.69} & --- & 36.94 & 31.78 & 10.91 & \multicolumn{1}{c|}{22.72} & --- & 29.49 & \textbf{35.08} \\
 & \multicolumn{1}{r|}{\texttt{RuAdapt Lit}} & \texttt{10,515} & 51.22 & \multicolumn{1}{c|}{41.64} & 49.94 & 53.74 & 48.54 & 22.66 & \multicolumn{1}{c|}{31.94} & 41.75 & \textbf{42.03} & 42.01 \\
 & \multicolumn{1}{r|}{\texttt{RuWikiLarge}} & \texttt{135,191} & 57.82 & \multicolumn{1}{c|}{44.38} & 55.03 & 51.97 & 40.82 & 24.24 & \multicolumn{1}{c|}{31.87} & 32.01 & 34.95 & \textbf{37.59} \\ \midrule
\multirow{2}{*}{\texttt{fr}} & \multicolumn{1}{r|}{\texttt{CLEAR}} & \texttt{3,179} & 55.00 & \multicolumn{1}{c|}{45.10} & 25.45 & 53.72 & 48.57 & 23.73 & \multicolumn{1}{c|}{32.17} & 34.86 & 30.85 & \textbf{35.37} \\
 & \multicolumn{1}{r|}{\texttt{WikiLargeFR}} & \texttt{148,276} & 58.51 & \multicolumn{1}{c|}{46.67} & 52.43 & 51.16 & 43.57 & 24.44 & \multicolumn{1}{c|}{32.23} & 35.20 & 38.22 & \textbf{39.23} \\ \midrule
\multirow{3.5}{*}{\texttt{en}} & \multicolumn{1}{r|}{\texttt{ASSET}} & \texttt{14,814} & 92.81 & \multicolumn{1}{c|}{88.11} & 88.26 & 81.20 & 85.90 & 20.73 & \multicolumn{1}{c|}{29.66} & 35.98 & \textbf{42.77} & 41.56 \\
 & \multicolumn{1}{r|}{\texttt{NewselaEN}} & \texttt{129,387} & 68.71 & \multicolumn{1}{c|}{52.30} & 62.78 & 51.51 & 55.68 & 26.17 & \multicolumn{1}{c|}{32.90} & 38.60 & \textbf{40.18} & 38.80 \\
 & \multicolumn{1}{r|}{\texttt{WikiAuto}} & \texttt{315,018} & 45.40 & \multicolumn{1}{c|}{41.31} & 37.95 & 35.30 & 36.91 & 20.93 & \multicolumn{1}{c|}{31.45} & 42.46 & \textbf{42.48} & 42.00 \\ \bottomrule
\end{tabular}%
}
\caption{\label{table:mt5-finetuning}
BLEU and SARI scores of mT5 fine-tuning experiments. Size refers to the total train sentence pairs after BLEU filtering. Best fine-tuned SARI score in bold. Results on training sets smaller than 3,000 pairs were omitted since this was not enough data to unlearn the pretraining objective.
}
\end{table*}

\section{Experiments}
\label{sec:Experiments}


\subsection{Evaluation Setup}

For automatic evaluation, we use SARI \citep{xu-etal-2016-optimizing}, the average of the F1 score for adding, keeping, and deleting n-grams ($n \in \{1,2,3,4\}$). SARI has been shown to correlate with human judgments of simplicity \citep{xu-etal-2016-optimizing}.  We also report BLEU \citep{papineni-etal-2002-bleu}, a common metric in machine translation. Although BLEU scores do not measure simplicity \citep{sulem-etal-2018-bleu}, we use them as a check for grammatically and meaning preservation \citep{xu-etal-2016-optimizing}.  We compute all evaluation metrics using \texttt{EASSE} evaluation suite \citep{alva-manchego-etal-2019-easse}.

\subsection{Baselines}

To put the results of our experiments in perspective, we compare them with two common baselines. \\

\noindent \textbf{Identity} The original sentence is copied and reported as the simplification.  This baseline earns high BLEU scores from the high token overlap between original and simple sentences. \\

\noindent \textbf{Truncation} The last 20\% of words are cut from the original sentence.  This baseline achieves high SARI scores because it balances keeping/deleting tokens, two operations SARI measures.

\subsection{Models}
For fine-tuning we used mT5 {\small Base} \citep{xue-etal-2021-mt5} (580M Parameters).  We calculated the S-BLEU score between the original and simple sentences in the training set and filtered out all sentences outside of the range [10,70] as done by \citet{maddela-etal-2021-controllable} to remove identical pairs (high BLEU) and misalignments (low BLEU). We also added four control tokens to the input sentence with information about the character-length compression, Levenshtein similarity, word rank, and dependency tree depth of the output following \citet{martin-etal-2020-controllable}. We used a grid search of control tokens on the dev set to find the combination that yielded the highest SARI for evaluation. We used BLOOM \citep{scao2022bloom} (176B Parameters) for a few-shot. We report hyperparameters, prompts, and details for both models in Appendix \ref{sec:experimental-details}.

\section{Results}
\label{sec:results}

\subsection{Fine-tuning Language Models}
\label{sec:fine-tuning-language-models}

We evaluated the mT5 models on all 27 datasets and fine-tuned them in 3 different settings. \texttt{Single}: On the training set of the dataset we are testing. \texttt{Language:} On the joint training set of all data in the same language. \texttt{All:} On the joint training set of all data across all languages. We remove results for training sets with fewer than 3,000 sentence pairs after S-BLEU filtering as we found this was not enough training data to unlearn the pre-training objective. We report the results of these experiments in Table \ref{table:mt5-finetuning}. \\


\noindent \textbf{Joint training improves performance in non-English languages.}
Joint all training improves SARI scores across every language besides English.  English already has a wealth of in-language data so it performs best with joint-language training.  There are specific datasets where joint-all does not achieve the highest SARI in other languages.  Typically these are within one SARI point.  Notably, PaCCSS-IT decreases in performance with more data.  This may be due to the automatic collection approach to PaCCSS-IT which is prone to collect slightly noisy data.  The similar BLEU scores to the identity baseline for all results suggests consistently high fluency.


\subsection{Zero-shot Cross-lingual Transfer}
\label{sec:cross-lingual}

We assess zero-shot cross-lingual and cross-domain transfer by training on one dataset and evaluating on another. We experiment with transfer to a small, domain-specific Italian dataset: Terence, and two low-resource language datasets: CBST in Basque and SimplifyUR in Urdu.  The transfer experiment results are shown in Table \ref{table:transfer}.\\


\begin{table}
\small
\setlength{\tabcolsep}{2pt}
\renewcommand{\arraystretch}{0.5}
\centering
\resizebox{0.385\textwidth}{!}{%
\begin{tabular}{@{}ccccrrr@{}}
\toprule
\multicolumn{7}{c}{Transfer to Italian: Terence \faIcon{book}} \\ \midrule
Scr & Fam & \multicolumn{1}{c|}{Lang} & Dom & \multicolumn{1}{r|}{Dataset} & \multicolumn{1}{c}{BLEU} & \multicolumn{1}{c}{SARI} \\ \midrule
 &  & \multicolumn{1}{c|}{} &  & \multicolumn{1}{r|}{Finetuned} & 7.96 & 23.92 \\ \midrule
 &  & \multicolumn{1}{c|}{} & \faIcon{book} & \multicolumn{1}{r|}{RuAdaptLit (ru)} & 57.07 & 35.19 \\
 &  & \multicolumn{1}{c|}{} & \faIcon{wikipedia-w} & \multicolumn{1}{r|}{RuWikiLarge (ru)} & 10.76 & 25.85 \\
\faIcon{check} &  & \multicolumn{1}{c|}{} & \faIcon{wikipedia-w} & \multicolumn{1}{r|}{WikiAuto (en)} & 21.90 & 30.89 \\
\faIcon{check} & \faIcon{check} & \multicolumn{1}{c|}{} & \faIcon{wikipedia-w} & \multicolumn{1}{r|}{WikiLargeFR (fr)} & 8.66 & 25.20 \\
\faIcon{check} & \faIcon{check} & \multicolumn{1}{c|}{\faIcon{check}} & \faIcon{wikipedia-w} & \multicolumn{1}{r|}{PaCCSS-IT (it)} & 40.59 & 36.99 \\ 
\midrule
\midrule
\multicolumn{7}{c}{Transfer to Basque: CBST \faIcon{flask}} \\ \midrule
Scr & Fam & \multicolumn{1}{c|}{Lang} & Dom & \multicolumn{1}{r|}{Dataset} & \multicolumn{1}{c}{BLEU} & \multicolumn{1}{c}{SARI} \\ \midrule
 &  & \multicolumn{1}{c|}{} &  & \multicolumn{1}{r|}{Finetuned} & 2.31 & 24.26 \\ \midrule
 &  & \multicolumn{1}{c|}{} & \faIcon[regular]{newspaper} \faIcon{globe} & \multicolumn{1}{r|}{EasyJA (ja)} & 1.87 & 26.67 \\
 &  & \multicolumn{1}{c|}{} & \faIcon{book} & \multicolumn{1}{r|}{RuAdaptLit (ru)} & 47.31 & 37.89 \\
\faIcon{check} &  & \multicolumn{1}{c|}{} & \faIcon[regular]{newspaper} & \multicolumn{1}{r|}{NewselaEN (en)} & 24.37 & 31.09 \\
\faIcon{check} &  & \multicolumn{1}{c|}{} & \faIcon{flask} \faIcon[regular]{newspaper} & \multicolumn{1}{r|}{PorSimples (pt-br)} & 7.97 & 29.44 \\ \midrule
\midrule
\multicolumn{7}{c}{Transfer to Urdu: SimplifyUR \faIcon[regular]{newspaper} \faIcon{book}} \\ \midrule
Scr & Fam & Lang & Dom & Dataset & \multicolumn{1}{c}{BLEU} & \multicolumn{1}{c}{SARI} \\ \midrule
 &  & \multicolumn{1}{c|}{} &  & \multicolumn{1}{r|}{Finetuned} & 5.20 & 33.18 \\ \midrule
 &  & \multicolumn{1}{c|}{} & \faIcon{wikipedia-w} & \multicolumn{1}{r|}{WikiAuto (en)} & 8.34 & 26.09 \\
 &  & \multicolumn{1}{c|}{} & \faIcon[regular]{newspaper} \faIcon{globe} & \multicolumn{1}{r|}{EasyJA (ja)} & 0.00 & 17.87 \\
 &  & \multicolumn{1}{c|}{} & \faIcon[regular]{newspaper} & \multicolumn{1}{r|}{NewselaEN (en)} & 0.25 & 18.03 \\
 &  & \multicolumn{1}{c|}{} & \faIcon{book} & \multicolumn{1}{r|}{RuAdaptLit (ru)} & 34.78 & 32.61 \\
 &  & \multicolumn{1}{c|}{} & \faIcon{flask} \faIcon[regular]{newspaper} & \multicolumn{1}{r|}{PorSimples (pt-br)} & 5.24 & 32.87 \\ \bottomrule
\end{tabular}
}
\caption{\label{table:transfer}
Transfer experiments to a domain specific small dataset (Terence) and two low resource language datasets (CBST, SimplifyUR).  We find that matching \underline{Scr}ipt, \underline{Fam}ily, \underline{\smash{Lang}}uage, and \underline{Dom}ain help improve transfer performance.
}
\end{table}

\noindent \textbf{Matching script and language improve transfer performance.}  In transfer experiments to Italian and Basque, we see a notable improvement in BLEU and SARI scores with datasets in matching scripts (Latin, in this case). The best transfer results in Italian come from another dataset in the same language, demonstrating that in-language transfer learning trumps cross-lingual transfer if the data is available. Transferring across scripts typically corresponds to lower performance except when domains match. \\

\noindent \textbf{Domain match can help regardless of script.}  In Urdu, we find that the best cross-lingual transfer results come from datasets in the same domain. This is true even though none of the transfer resources are in the Arabic script. In the transfer to Terence (Italian literature corpus), the Russian dataset with the matching domain, RuAdaptLit, outperforms RuWikiLarge, another Russian dataset from a different domain. Still, the domain alone does not guarantee strong transfer performance. EasyJA and NewselaEN performed poorly in Urdu transfer despite matching in domain.

\noindent \textbf{Russian is a good candidate language for cross-lingual transfer.}  For every test setting, the Russian corpus, RuAdaptLit, transfers well to the target dataset. Additionally, RuWikiLarge transfers better to Terence than the comparable WikiLargeFR, even though both datasets are machine translations of the same English WikiLarge corpus. This suggests that Russian is a good candidate language for cross-lingual transfer, which is in line with the findings of \citet{turc2021revisiting} that  Russian is a better choice than English as a pivot language for zero-shot cross-lingual transfer.

\subsection{Prompting Multilingual Language Models}
\label{sec:prompting-mlms}

We assess few-shot performance in two settings. \texttt{Semantic Similarity}:  We computed LASER sentence embeddings \citep{schwenk2017learning} for all sentences in the train, test, and validation set. During evaluation, we used the $k$ nearest neighbors in the train set by cosine distance as examples. \texttt{Random Sampling}: We choose $k$ random sentences from the train set as examples during evaluation. We highlight interesting findings of our few-shot experiments here. Fewshot results for all datasets are available in Table \ref{table:fewshot}. \\

\begin{figure}[t!]
    \centering
    \includegraphics[width=0.99\linewidth]{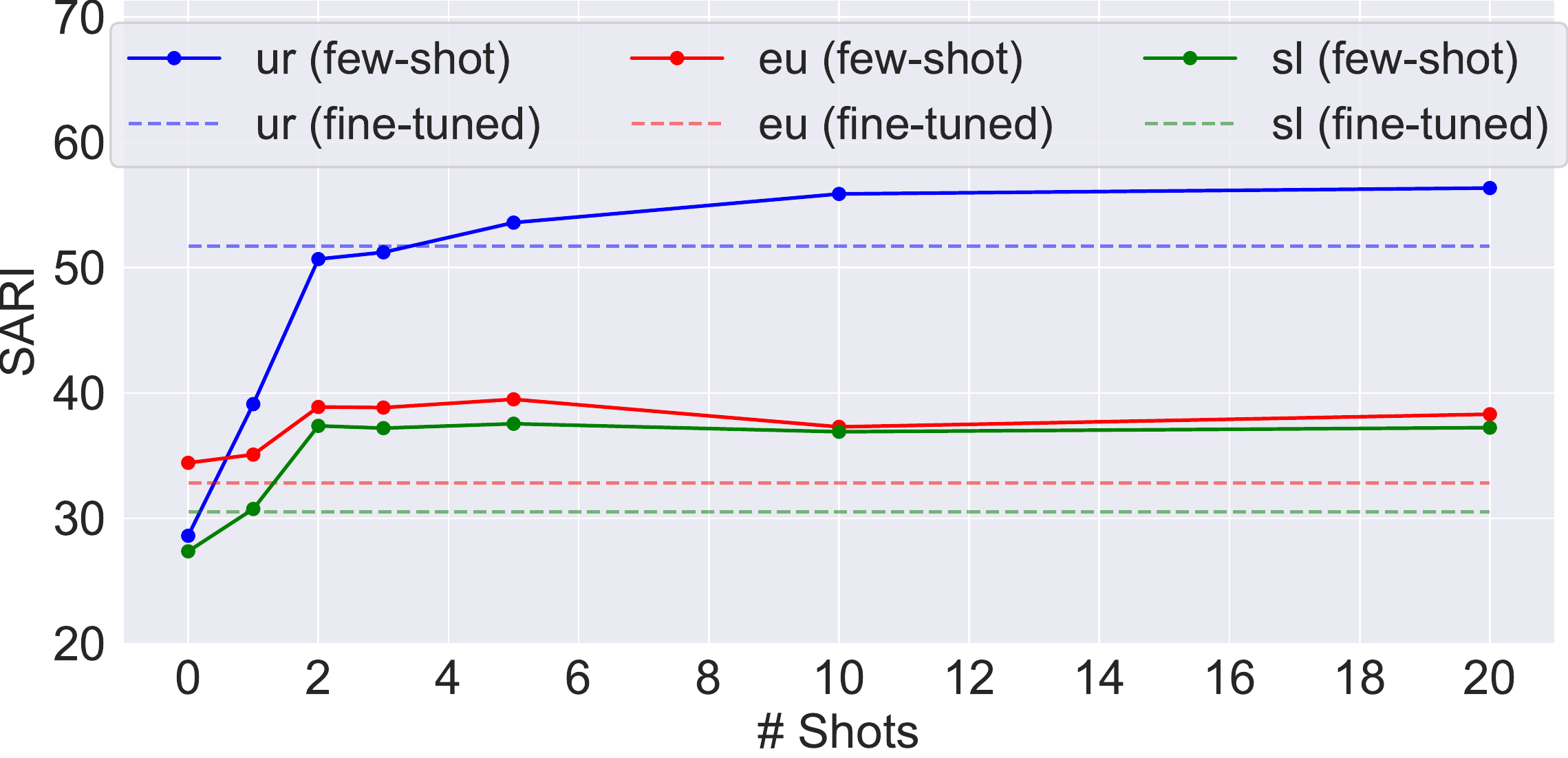}
    \caption{Semantic similarity fewshot performance in low-resource languages. Fewshot prompting achieves higher SARI than mt5 finetuned.}
    \label{fig:low-resource-fewshot}
\end{figure}

\begin{figure}[t!]
    \centering
    \includegraphics[width=0.99\linewidth]{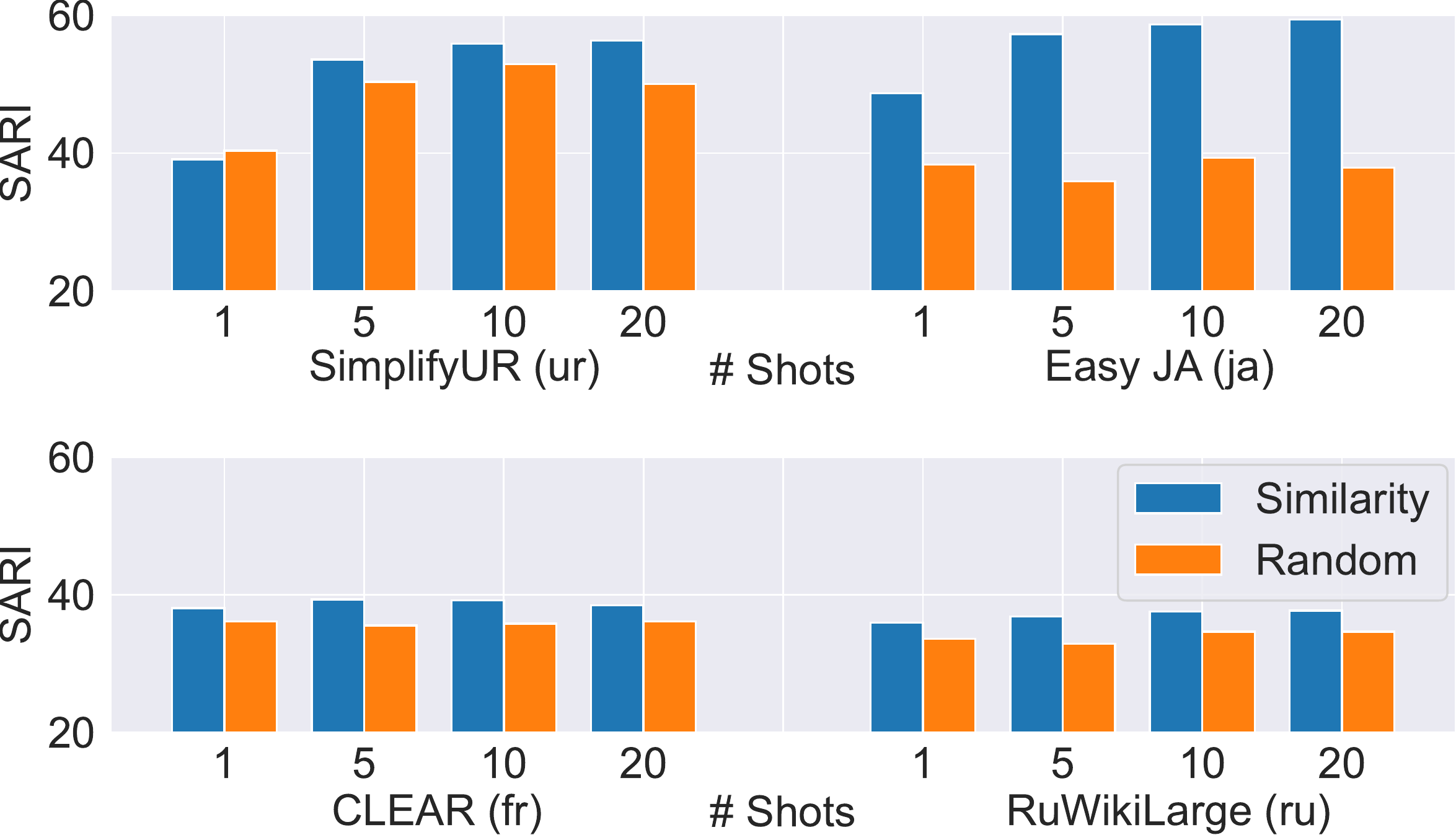}
    \caption{Semantic similarity vs random sampling fewshot performance on four diverse datasets.  Semantic similarity consistently scores above random sampling.}
    \label{fig:fewshot-bar}
\end{figure}

\begin{table*}[]
\small
\setlength{\tabcolsep}{2pt}
\renewcommand{\arraystretch}{1.0}
\centering
\resizebox{0.99\textwidth}{!}{%
\begin{tabular}{@{}rllllllllllll@{}}
\toprule
\multicolumn{1}{l}{} & \multicolumn{3}{c}{\textbf{English (ASSET)}} & \multicolumn{3}{c}{\textbf{Russian (RuAdaptLit)}} & \multicolumn{3}{c}{\textbf{Italian (Terence)}} & \multicolumn{3}{c}{\textbf{Urdu (SimplifyUR)}} \\ \midrule
\multicolumn{1}{l|}{} & \multicolumn{1}{c}{Adequacy} & \multicolumn{1}{c}{Fluency} & \multicolumn{1}{c|}{Simplicity} & \multicolumn{1}{c}{Adequacy} & \multicolumn{1}{c}{Fluency} & \multicolumn{1}{c|}{Simplicity} & \multicolumn{1}{c}{Adequacy} & \multicolumn{1}{c}{Fluency} & \multicolumn{1}{c|}{Simplicity} & \multicolumn{1}{c}{Adequacy} & \multicolumn{1}{c}{Fluency} & \multicolumn{1}{c}{Simplicity} \\ \midrule
\multicolumn{1}{r|}{Reference} & 4.60{\tiny $\pm0.22$} & 4.85{\tiny $\pm0.11$} & \multicolumn{1}{l|}{4.13{\tiny $\pm0.26$}} & 3.70{\tiny$\pm0.47$} & 4.45{\tiny$\pm0.32$} & \multicolumn{1}{l|}{2.50{\tiny$\pm0.24$}} & 4.73{\tiny$\pm0.55$} & 4.88{\tiny$\pm0.33$} & \multicolumn{1}{l|}{2.98{\tiny$\pm1.37$}} & 4.83{\tiny$\pm0.38$} & 5.00{\tiny$\pm0.00$} & 4.25{\tiny$\pm1.17$} \\ \midrule
\multicolumn{1}{r|}{mT5 Single} & 4.45{\tiny $\pm0.25$} & 4.95{\tiny $\pm0.07$} & \multicolumn{1}{l|}{3.00{\tiny $\pm0.34$}} & 4.50{\tiny$\pm0.31$} & 4.78{\tiny$\pm0.19$} & \multicolumn{1}{l|}{2.25{\tiny$\pm0.15$}} & 1.23{\tiny$\pm0.73$} & 1.15{\tiny$\pm0.66$} & \multicolumn{1}{l|}{1.15{\tiny$\pm0.66$}} & 2.38{\tiny$\pm1.37$} & 2.10{\tiny$\pm1.32$} & 1.10{\tiny$\pm0.30$} \\
\multicolumn{1}{r|}{mT5 Joint Language} & 4.65{\tiny $\pm0.19$} & 4.98{\tiny $\pm0.05$} & \multicolumn{1}{l|}{3.38{\tiny $\pm0.25$}} & 4.78{\tiny$\pm0.26$} & 5.00{\tiny$\pm0.00$} & \multicolumn{1}{l|}{2.48{\tiny$\pm0.24$}} & 4.78{\tiny$\pm0.48$} & 4.83{\tiny$\pm0.55$} & \multicolumn{1}{l|}{2.55{\tiny$\pm1.01$}} & \multicolumn{1}{c}{—} & \multicolumn{1}{c}{—} & \multicolumn{1}{c}{—} \\
\multicolumn{1}{r|}{mT5 Joint All} & 4.64{\tiny $\pm0.18$} & 4.93{\tiny $\pm0.08$} & \multicolumn{1}{l|}{2.94{\tiny $\pm0.21$}} & 4.23{\tiny$\pm0.38$} & 4.85{\tiny$\pm0.21$} & \multicolumn{1}{l|}{2.75{\tiny$\pm0.32$}} & 4.18{\tiny$\pm1.20$} & 4.65{\tiny$\pm0.70$} & \multicolumn{1}{l|}{2.50{\tiny$\pm1.09$}} & 4.25{\tiny$\pm1.13$} & 4.88{\tiny$\pm0.65$} & 2.95{\tiny$\pm1.30$} \\ \midrule
\multicolumn{1}{r|}{mT5 English Transfer} & \multicolumn{1}{c}{—} & \multicolumn{1}{c}{—} & \multicolumn{1}{c|}{—} & 2.18{\tiny$\pm0.49$} & 1.70{\tiny$\pm0.40$} & \multicolumn{1}{l|}{1.18{\tiny$\pm0.12$}} & 1.73{\tiny$\pm1.45$} & 1.88{\tiny$\pm1.54$} & \multicolumn{1}{l|}{1.25{\tiny$\pm0.59$}} & 1.83{\tiny$\pm1.08$} & 1.40{\tiny$\pm0.63$} & 1.00{\tiny$\pm0.00$} \\
\multicolumn{1}{r|}{mT5 Russian Transfer} & 4.25{\tiny $\pm0.33$} & 3.93{\tiny $\pm0.16$} & \multicolumn{1}{l|}{2.63{\tiny $\pm0.30$}} & \multicolumn{1}{c}{—} & \multicolumn{1}{c}{—} & \multicolumn{1}{c|}{—} & 4.53{\tiny$\pm1.01$} & 4.70{\tiny$\pm0.72$} & \multicolumn{1}{l|}{2.35{\tiny$\pm0.92$}} & 3.70{\tiny$\pm1.73$} & 3.60{\tiny$\pm1.85$} & 1.60{\tiny$\pm0.50$} \\ \midrule
\multicolumn{1}{r|}{BLOOM 5 Shot (Rand)} & 4.63{\tiny $\pm0.25$} & 4.75{\tiny $\pm0.18$} & \multicolumn{1}{l|}{3.10{\tiny $\pm0.40$}} & 4.65{\tiny$\pm0.27$} & 4.78{\tiny$\pm0.22$} & \multicolumn{1}{l|}{2.33{\tiny$\pm0.20$}} & 4.33{\tiny$\pm1.02$} & 4.53{\tiny$\pm0.85$} & \multicolumn{1}{l|}{2.45{\tiny$\pm1.11$}} & 4.95{\tiny$\pm0.22$} & 4.93{\tiny$\pm0.35$} & 3.28{\tiny$\pm1.69$} \\
\multicolumn{1}{r|}{BLOOM 5 Shot (Sim)} & 4.63{\tiny $\pm0.22$} & 4.80{\tiny $\pm0.13$} & \multicolumn{1}{l|}{2.88{\tiny $\pm0.32$}} & 4.63{\tiny$\pm0.28$} & 4.95{\tiny$\pm0.07$} & \multicolumn{1}{l|}{2.43{\tiny$\pm0.24$}} & 4.00{\tiny$\pm1.47$} & 4.58{\tiny$\pm1.01$} & \multicolumn{1}{l|}{2.38{\tiny$\pm1.19$}} & 4.83{\tiny$\pm0.68$} & 4.85{\tiny$\pm0.70$} & 3.28{\tiny$\pm1.71$} \\ \bottomrule
\end{tabular}
}
\caption{\label{table:manual-eval}
Human evaluation adequacy, fluency, and simplicity scores for mT5 fine-tuned, mt5 zero-shot cross-lingual transfer, and BLOOM few-shot in English, Russian, Italian, and Urdu.  Scores are averaged over 20 ratings per system with 95\% confidence intervals.
}
\end{table*}

\noindent \textbf{Few-shot prompting is promising for low-resource languages.} Figure \ref{fig:low-resource-fewshot} shows semantic similarity sampling few-shot performance for the three low-resource languages in our study: Urdu, Basque, and Slovene. In all low-resource languages tested, few-shot outperforms fine-tuned mT5 trained on all data (+4.62, +5.48, +6.72 SARI, respectively).  Within five examples few-shot exceeds fine-tuned performance.  With limited resources, few-shot prompting is a good alternative to fine-tuning. \\

\noindent \textbf{Semantic similarity outperforms random sampling.}  Figure \ref{fig:fewshot-bar} shows semantic similarity vs. random sampling for few-shot evaluation on four diverse datasets: SimplifyUR (low resource language), EasyJapanese (manually simplified), CLEAR (medical domain), and RuWikiLarge (machine translated). In all cases prompting with semantic search outperformed random examples. This trend persists across languages and domains. Typically more examples improved performance, but any past five had marginal benefits.

\section{Human Evaluation}
\label{sec:manual-analysis}

For manual evaluation, we enlisted eight volunteers to annotate system outputs in English, Russian, Italian, and Urdu (two in each language) for three properties: Adequacy (is the meaning preserved?), Fluency (is the simplification eloquent/grammatical?), and Simplicity (is the output simpler?). This follows the standard annotation methodology in text simplification research \citep{martin-etal-2022-muss, xu-etal-2016-optimizing}. We asked annotators to rate 20 sentences for each model using a Likert scale from 1-5. We report the manual evaluation key in Table \ref{table:manual-eval-instructions}.  Since our ratings are ordinal data, we measure annotator agreement using Krippendorff's alpha \cite{krippendorff2011computing} through the Fast Krippendorff Library \cite{castro-2017-fast-krippendorff}.  We achieve a high-reliability coefficient in all languages suggesting good annotator agreement.  Specifically we calculate $\alpha = 0.80$ in English, $\alpha = 0.79$ in Russian, $\alpha = 0.75$ in Italian, and $\alpha = 0.86$ in Urdu.

Table \ref{table:manual-eval} shows the manual evaluation results. Single dataset fine-tuned performance was deficient in Italian and Urdu because these datasets were small resources (1,012 pairs and 736 pairs, respectively).  The Joint-all model performed consistently well across all datasets but did not outperform English language training for English.  This aligned with our earlier findings (\S \ref{sec:fine-tuning-language-models}) and suggests that very large-scale in-language data is better than multilingual training if such data is available.  Russian transfer using RuAdaptLit outperformed English transfer to both Italian and Urdu, reinforcing our observation that Russian is a strong choice for cross-lingual transfer (\S \ref{sec:cross-lingual}). We observe the best model scores from five-shot BLOOM to be on par with the reference simplifications in Italian/Russian and scoring slightly below reference simplifications in English/Urdu. This finding suggests that Few-shot prompting is effective for text simplification in both high and low-resource languages.  In the low-resource Urdu setting, few-shot prompting yielded the best results, further substantiating our observations from few-shot prompting experiments (\S \ref{sec:prompting-mlms}).

\section{Conclusion}


We release {\sc MultiSim}, the first multilingual text simplification benchmark, a collection of 27 sentence-aligned parallel corpora in 12 diverse languages. We collected these resources by surveying the literature for all existing text simplification resources in non-English languages, which were created via distinct methodologies that we categorize into four main approaches. Using {\sc MultiSim}, we perform fine-tuning, few-shot, and zero-shot cross-lingual transfer experiments with generative multilingual language models (mT5, BLOOM), which revealed new insights in multilingual text simplification. Our results demonstrate the value of domain and script match for zero-shot cross-lingual transfer. We show that Russian is a good candidate pivot language, outperforming transfer from English in two of our case studies on low-resource and out-of-script languages. Further, we show that few-shot prompting BLOOM with examples obtained via semantic similarity outperforms fine-tuned models for low-resource languages. By releasing this benchmark, we hope to encourage and enable the development and evaluation of multilingual models and evaluation metrics for text simplification.

\section*{Limitations}
This benchmark compiled and analyzed existing resources collected from diverse methods and domains. Although we demonstrated how careful use of these resources could transfer well to other resources, along with a manual analysis of a varied set of corpora, we cannot guarantee the quality of each resource or validate the methods that the original authors used to create them. We explore each dataset's linguistic properties in Appendix \ref{sec:analysis}. However, we encourage a deeper exploration of the quality of individual resources by researchers that speak the 12 languages included in this benchmark and corresponding data loaders.

Additionally, the human evaluation performed in this study was limited in scope and served primarily to validate the findings by automatic metrics.  A more extensive evaluation with more annotators evaluating more sentences would be beneficial in order to draw further conclusions.

Furthermore, some of the resources discussed in this paper were automatically aligned. Although Neural CRF models in English have been shown to yield high-quality alignments \citep{jiang-etal-2020-neural}, other alignment algorithms such as TF-IDF scoring \citep{nelken-shieber-2006-towards} have been shown to result in a high number of false positives \citep{xu-etal-2015-problems}. Future work could include realigning automatically aligned corpora using an embedding-based sentence alignment model trained on manually annotated alignment data \citep{jiang-etal-2020-neural}.  We will continue updating this benchmark as updates are made to the underlying datasets, and new multilingual resources are released.

\section*{Acknowledgments}
We thank Yang Chen and Yao Dou as well as three anonymous reviewers for their helpful feedback on this work. We also thank Govind Ramesh, Nour Allah El Senary, Luca Castagna, Lory O'Brien, Franco Paglione, Livia Paglione, Anton Lavrouk, Leah Levin, Irina Levin, Muhammad Hassan Maqsood, and Talha Ahmad Khan for their help with human evaluation. Furthermore, we thank Mounica Madella for sharing her control token scripts. This research is supported in part by the NSF awards IIS-2144493 and IIS-2112633, ODNI and IARPA via the BETTER program (contract 2019-19051600004) and the HIATUS program (contract 2022-22072200004). The views and conclusions contained herein are those of the authors and should not be interpreted as necessarily representing the official policies, either expressed or implied, of NSF, ODNI, IARPA, or the U.S. Government. The U.S. Government is authorized to reproduce and distribute reprints for governmental purposes notwithstanding any copyright annotation therein.

\bibliography{anthology,custom}
\bibliographystyle{acl_natbib}

\newpage
\clearpage

\appendix

\section{Resource Summary}
\label{sec:summary}

Here we provide a brief summary of the 34 monolingual text simplification parallel corpora surveyed in this work.  Our summary focuses on the domain, target audience, and collection strategy of each resource.

\subsection{English}
\textbf{ASSET} \citep{alva-manchego-etal-2020-asset} is a high quality collection of 2,390 original sentences from the TurkCorpus \citep{xu-etal-2016-optimizing} which sampled from Wikipedia.  ASSET contains 10 manually written simplifications for each of the original sentences using a variety of rewrite operations.

The \textbf{Newsela English} corpus \citep{xu-etal-2015-problems} was produced by professional writers at Newsela, a U.S. company dedicated to providing high-quality simplifications of informational content for schools.  Each article was written at 5 levels, including the original article (level 0) and 4 levels of simplification.  For this paper we use the neural CRF alignments provided by \citet{jiang-etal-2020-neural}.

\textbf{WikiAuto} \citep{jiang-etal-2020-neural} is a neural CRF aligned corpus of original and simple wikipedia documents.  This is currently the largest text simplification resource available with over ten million original sentences.  Once aligned, the number of sentence pairs reduces to just under 600,000.  This is because most of the original and simple wikipedia articles are not exact rewrites.

\subsection{Spanish}
The \textbf{FIRST} corpus \citep{orasan2013text} was collected as a part of the EU-funded FIRST project to develop a tool to assist people with autism spectrum disorder (ASD) in reading and understanding written documents. The corpus contains 25 original and simplified documents from literature, news, health, general culture, and instructions. The simplifications were performed manually by experts with experience working with individuals with ASD.

The \textbf{Simplext} corpus \citep{10.1145/2738046} consists of 193 articles from the news outlet, Servimedia, which were manually simplified based on specific simplification recommendations \citep{anula2011pautas}. The articles span four domains: national news, international news, society, and culture.

The \textbf{Newsela Spanish} corpus \citep{xu-etal-2015-problems} was created alongside the Newsela English corpus for the same audience of students.  For analysis in this work the corpus was aligned between adjacent levels (i.e., 0-1, 1-2, etc.) using CATS aligner \citep{stajner-etal-2017-sentence}.  

\subsection{Italian}

The \textbf{Terence and Teacher} corpora \citep{brunato-etal-2015-design} were the first two Italian parallel corpora for ATS. The \textbf{Terence} corpus consists of 32 simplified short stories for children and the \textbf{Teacher} corpus consists of 18 texts from educational websites. The Terence corpus was simplified by experts with target rules, while the Teacher corpus was simplified by teachers targeting second language learners.

\textbf{SIMPITIKI} \citep{Tonelli2016} is comprised of a Wikipedia-based sub corpus and a government-document-based sub corpus.   \textbf{SIMPITIKI Wiki} used crowdsourced Wikipedia simplifications by extracting edits with keywords such as "simplified" from the edit history of an Italian Wikipedia dump.  Prior work has shown that the quality of Wikipedia simplifications is not guaranteed \citep{xu-etal-2015-problems}, however, the simplifications were manually selected to ensure quality.  This is the primary corpus studied as Simpitiki throughout this paper.  SIMPITIKI PA was produced by the authors for comparing simplification operations and was later absorbed as a subset of the \textbf{AdminIT} corpus \citep{miliani-etal-2022-neural}.  The authors manually simplified public administration documents about building permits and kindergarten admittance.

\textbf{PaCCSS-IT} \citep{brunato-etal-2016-paccss} instead extracted simplification pairs from a large corpus of text. The researchers assumed that in a large enough text dataset (i.e., scraped from the internet), both complex and simple sentences that have similar meanings were bound to exist. \citet{brunato-etal-2016-paccss} found over 63,000 such matches using cosine similarity and an SVM using lexical, morpho-syntactic, and syntactic features. Upon manual analysis by the authors, it turned out that about 85\% of the pairs were aligned correctly, while 74\% of those correct pairs were actually simplifications.

\subsection{French}
\textbf{CLEAR} \citep{cardon-grabar-2019-parallel} is a parallel corpus of biomedical texts written in French and automatically aligned using a random forests classifier of 10 textual features.  In a manual assessment of 30 documents, the authors found that 98.75\% of alignments were correct suggesting a high-precision sentence alignment.

\textbf{WikiLargeFR} \citep{cardon-grabar-2020-french} is a machine-translated version of the English WikiLarge corpus \citep{zhang-lapata-2017-sentence} using OpenNMT-py \citep{klein-etal-2017-opennmt}.  It has 297,753 sentence pairs but different exact counts of complex and simple sentences when accounting for sentence splitting.  It was created as a comparison with CLEAR for biomedical text simplification.

\textbf{Alector} \citep{gala-etal-2020-alector} contains expert simplified versions of 79 texts selected at a 2-4th grade reading level.  The authors showed that simplifying the texts reduced misreadings in dyslexic and low literacy readers.

\subsection{Japanese}
\textbf{Japanese News} \citep{goto-etal-2015-japanese} is created from Japan Broadcasting Corporation (NHK)'s online service NEWS WEB EASY which provides original news articles rewritten by Japanese instructors for simplicity.  \citet{goto-etal-2015-japanese} used a dynamic programming aligner to align 10,651 sentences.  They also manually aligned 2,735 sentences. 

\textbf{EasyJapanese} \citep{maruyama-yamamoto-2018-simplified} was created for the purpose of improving Japanese resources for foreign citizens. Since Japanese language learners usually have a limited vocabulary, the authors decided to produce a parallel corpus using a vocabulary of 2,000 common Japanese words. 5 students in the lab manually simplified 50,000 sentences with S-BLEU \citep{papineni-etal-2002-bleu} scores between simplifications of the same sentence ranging from 0.58 to 0.63. The corpus was built from a previous bilingual web crawl of Japanese and English news articles called the Tanaka corpus \citep{tanaka2001compilation}.

\textbf{EasyJapaneseExtended} \citep{katsuta-yamamoto-2018-crowdsourced} included 34,400 more sentences from the Tanaka corpus \citep{tanaka2001compilation} with simplifications crowdsourced from the CrowdWorks\footnote{https://crowdworks.jp/} platform. The authors measured the S-BLEU scores on 100 sentences that each of the 7 workers simplified and found that for 70\% of the workers, S-BLEU scores exceeded 0.4.

\subsection{Brazilian Portuguese}
\textbf{PorSimples} \citep{aluisio-gasperin-2010-fostering} was one of the first ATS projects.  It had the express purpose of simplifying texts for individuals with reading difficulties.  For this project, \citet{caseli2009building} collected a parallel corpus of 104 news articles.  The articles were simplified at 2 levels by a linguist specializing in text simplification.  In ``natural'' simplifications, the linguist could choose how to simplify the text.  In ``strong'' simplifications, the linguist had to follow very specific rules \citep{specia2008manual,10.1145/1456536.1456540}.

\subsection{German}
\textbf{Simple German} \citep{battisti-etal-2020-corpus} started from \citet{klaper-etal-2013-building}'s work that scraped texts from the internet and aligned them using a monolingual sentence alignment algorithm of \citet{barzilay-elhadad-2003-sentence}. \citet{battisti-etal-2020-corpus} further improved upon this with much more data and better alignment algorithms of CATS \citep{stajner-etal-2018-cats} and MASSAlign \citep{paetzold-etal-2017-massalign}. The original paper reports 378 documents with 17,121 original and 21,072 simple sentences. Note that, these numbers differ from those in Table \ref{table:corpora-stats} as the availability of online articles has changed since the original publication.



\textbf{TextComplexityDE} \citep{https://doi.org/10.48550/arxiv.1904.07733} was created to measure text complexity in German.  1000 sentences were taken from German Wikipedia and 100 sentences from Simple German \citep{klaper-etal-2013-building}.  German second language learners rated the sentences on a 7-point Likert scale for complexity. The 250 most complex sentences were manually simplified by native speakers.

\textbf{GEOLinoTest} \citep{mallinson-etal-2020-zero} was built as an evaluation dataset for a zero-shot ATS model. \citet{mallinson-etal-2020-zero} extracted 20 articles about nature, physics, and people from GeoLino, a children's magazine.  A German linguist simplified them to a five to seven-year-old reading level.

\textbf{German News} \citep{sauberli-etal-2020-benchmarking} contains 3,616 sentences simplified by the Austrian Press Agency on politics, economy, culture, and sports.  The authors trained neural simplification models on the corpus and found success, including a simple sentence matched to itself during training \citep{palmero-aprosio-etal-2019-neural}.

The \textbf{Klexikon} corpus \citep{aumiller-gertz-2022-klexikon} is a mapping of documents from German Wikipedia to the children's encyclopedia site: Klexikon \footnote{\url{https://klexikon.zum.de}}.  Klexikon targets German readers aged six to twelve.  Like WikiAutoEN, Klexikon is a large-scale alignment of documents, however, unlike similar resources, Klexikon does not yet have a gold-standard automatic sentence alignment.  The authors of Klexikon are working to release this alignment which will make Klexikon a very large German text simplification resource.

\textbf{Simple Patho} \citep{trienespatient} is an upcoming biomedical text simplification corpus for German of 851 clinical reports simplified by nine medical students.  Due to privacy concerns, the dataset is not yet available.  When it is released, it will serve as a large and high-quality medical text simplification corpus for the community.

\subsection{Basque}
\textbf{The Corpus of Basque Simplified Texts (CBST)} \citep{10.1007/s10579-017-9407-6} contains 227 sentences from 3 science popularization documents simplified to two distinct levels. Two people simplified the documents: a translator without simplification experience who focused on simplification guidelines \citep{mitkov-stajner-2014-fewer} and a language teacher who focused on intuitive transformations.

\subsection{Danish}
\textbf{DSim} \citep{klerke-sogaard-2012-dsim} extracted 3,701 pairs of news telegrams from the Danish Broadcasting Corporation's news and educational services.  The corpus was simplified by journalists to help reading-impaired adults and adult learners of Danish.  The documents were automatically sentence-aligned using TF-IDF scores.

\subsection{Urdu}
\textbf{SimplifyUREval} \citep{qasmi-etal-2020-simplifyur} was a corpus made for evaluating the ATS model SimplifyUR.  The model used word substitutions to propose simplifications  to Urdu.  The evaluation corpus contains 500 sentences from newspapers, magazines, books, and literary journals manually simplified by a linguist with a doctorate in Urdu.  Two additional native Urdu speakers manually verified 50 sentences and had an inter-annotator agreement of 0.9, measured by Cohen's Kappa.

\subsection{Russian}
\textbf{RuWikiLarge} \citep{sakhovskiy2021rusimplesenteval} is a machine translated version of EnWikiLarge \citep{zhang-lapata-2017-sentence}.  It has 248,111 sentence pairs but different exact counts of complex and simple sentences when accounting for sentence splitting.  It was created as a resource for the RuSimpleSentEval shared task \citep{sakhovskiy2021rusimplesenteval}.

\textbf{RuSimpleSentEval (RSSE)} \citep{sakhovskiy2021rusimplesenteval} was mined from Russian Wikipedia's most popular pages also for the RuSimpleSentEval.  Crowdsource workers on Yandex Toloka were asked to simplify the sentences.

\textbf{RuAdapt} \citep{dmitrieva-tiedemann-2021-creating} was created from 6 collections of several novels each and 16 individual classical and modern Russian literature books.  The simplified books were prepared by Russian-as-a-Foreign-Language (RaaFL) teachers.  The corpus was aligned with Bleualign \citep{sennrich-volk-2010-mt} and CATS \citep{stajner-etal-2017-sentence}.  In addition, researchers contributed both encyclopedic simplifications and fairytale simplifications \citep{Dmitrieva2021Quantitative}.

\subsection{Slovene}
The \textbf{SloTS} corpus \citep{11356/1682} pulls from 10 existing texts simplified by the RISA Institute.  The RISA Institute is an organization that publishes easy-to-read Slovenian novels and news.  SloTS is a collection of about 1,000 sentence pairs sampled from 10 novels and manually aligned between the original and simplified versions.  This dataset was used to train a Slovene text simplification model built on SloT5 \citep{ulvcar2022sequence}.

\subsection{Arabic}
\textbf{Saaq al-Bambuu} \citep{al2016saud} is an internationally acclaimed Arabic novel that has been rewritten for Arabic-as-a-second-language learners.  \citet{khallaf2022towards} sampled 2,980 parallel sentences from the original and simplified books at two different levels.  Unfortunately, due to copyright restrictions, the corpus is not available publicly.

\section{Corpus Analysis}
\label{sec:analysis}
This section provides some key statistics of the corpora introduced in Section \ref{sec:datasets}.  In performing this analysis, we hope to highlight the differences between various corpora and offer greater insight into their quality and composition to facilitate future research.

\subsection{Basic Statistics}
\label{sec:appendix-basic-stats}
The Basic Statistics computed were vocab size, token count, average tokens per sentence, average characters per token, and average sentences per doc.  All of the statistics besides average sentences per document depend heavily on the word tokenization of the corpus.  For space-delimited languages, we used the Toktok tokenizer \citep{Dehdari2014ANS} from the natural language toolkit (NLTK) \citep{bird2009natural}.  For our work, this included all languages besides Urdu and Japanese.  For Urdu tokenization, we used UrduHack\footnote{\url{https://docs.urduhack.com/en/stable/}}, a Python library built for academic researchers and professional developers working on Urdu NLP projects.  For Japanese, we used fugashi \citep{mccann-2020-fugashi}, a tool for Japanese tokenization in Python.  We used the unidic dictionary \citep{den-etal-2008-proper} to define the Japanese vocabulary for tokenization.

Basic statistics results are reported in Table \ref{table:basic-stats}.  In general, the trend between original and simplified texts was reduced vocab size, reduced token count, reduced sentence length, and reduced word length.  This corresponds well with the expected simplification operations of replacing longer, more complicated words with shorter, more common words.  It also aligns with a common simplification strategy of splitting longer sentences into two or more short sentences.  This explains why the average sentence length decreased but in many document-aligned corpora, the average sentences per document increased.  There were some exceptions to this.  Notably, the Japanese corpora had a higher token count and higher average tokens/sentence.  This could be due to the limited vocabulary used when creating these two corpora.  As a part of the Easy Japanese corpus creation authors were limited to just 2,000 Japanese words.  The fugashi tokenizer identified more than 2,000, but still reported a large drop in vocab size.  The authors may have needed to be creative with how they chose to rewrite sentences using the limited vocabulary.  This could've led to many edits where the author explained a complex idea in several simple words instead of using one more complicated word.  Another easily explained set of outliers is SimplifyUR and RSSE having larger vocab sizes from simple to complex.  Both of these corpora allow multiple translations of the same original sentence.  This means for a given sentence pair with the same original sentence the original vocab size will remain the same while the simple vocab size might increase.

\begin{table*}
\tiny
\setlength{\tabcolsep}{2pt}
\renewcommand{\arraystretch}{0.9}
\centering
\resizebox{0.99\textwidth}{!}{%
\begin{tabular}{@{}l|r|cc|cc|cc|cc|cc@{}}
\toprule
\textbf{Corpus} & \textbf{Lang} & \multicolumn{2}{c|}{\textbf{Vocab Size}} & \multicolumn{2}{c|}{\textbf{Token Count}} & \multicolumn{2}{c|}{\textbf{Avg Tok/Sent}} & \multicolumn{2}{c|}{\textbf{Avg Char/Tok}} & \multicolumn{2}{c}{\textbf{Avg Sent/Doc}} \\
 & \multicolumn{1}{l|}{} & \textbf{orig $\uparrow$} & \textbf{simp $\downarrow$} & \textbf{orig $\uparrow$} & \textbf{simp $\downarrow$} & \textbf{orig $\uparrow$} & \textbf{simp $\downarrow$} & \textbf{orig $\uparrow$} & \textbf{simp $\downarrow$} & \textbf{orig} & \textbf{simp} \\ \midrule
DSim & da & 57,308 & 40,220 & 953,201 & 796,201 & 19.91 & 13.15 & 5.57 & 5.36 & --- & --- \\
GEOLino & de & 4,467 & 4,266 & 19,185 & 17,889 & 16.01 & 14.93 & \textbf{5.68} & \textbf{5.74} & --- & --- \\
German News A2 & de & 23,542 & 7,764 & 147,905 & 78,946 & 20.30 & 11.23 & 6.40 & 5.79 & 4.03 & 3.89 \\
German News B2 & de & 25,039 & 10,473 & 160,188 & 93,283 & 20.14 & 12.75 & 6.43 & 5.99 & 4.96 & 4.56 \\
Klexikon & de & 706,243 & 55,868 & 15,240,505 & 1,239,694 & 19.77 & 12.80 & 6.42 & 5.53 & 266.53 & 33.48 \\
Simple German & de & 35,763 & 18,753 & 313,622 & 199,861 & 24.50 & 23.79 & 6.60 & 6.38 & 57.16 & 37.50 \\
TextComplexityDE & de & 3,068 & 2,760 & 7,485 & 7,092 & 29.94 & 28.37 & 6.78 & 6.62 & 10.87 & 10.87 \\
ASSET & en & \textbf{11,998} & \textbf{19,320} & 521,940 & 448,376 & 22.13 & 19.01 & 5.28 & 5.18 & --- & --- \\
Newsela EN 0-1 & en & 68,972 & 61,115 & 2,187,046 & 1,881,631 & 23.95 & 19.83 & 5.08 & 5.06 & 48.52 & 50.41 \\
Newsela EN 1-2 & en & 61,115 & 53,673 & 1,881,631 & 1,733,011 & 19.83 & 16.84 & 5.06 & 4.98 & 50.41 & 54.67 \\
Newsela EN 2-3 & en & 53,673 & 42,879 & 1,733,011 & 1,458,744 & 16.84 & 13.93 & 4.98 & 4.86 & 54.67 & 55.65 \\
Newsela EN 3-4 & en & 42,879 & 34,104 & 1,458,744 & 1,144,534 & 13.93 & 11.48 & 4.86 & 4.75 & 55.65 & 53.00 \\
WikiAuto & en & 2,009,681 & 419,496 & 265,352,569 & 22,170,411 & 26.16 & 17.86 & 5.19 & 4.96 & --- & --- \\
Newsela ES 0-1 & es & 27,950 & 23,452 & 323,034 & 257,905 & 28.28 & 23.31 & 5.37 & 5.36 & 47.01 & 45.52 \\
Newsela ES 1-2 & es & 23,452 & 20,582 & 257,905 & 225,659 & 23.31 & 19.35 & 5.36 & 5.31 & 45.52 & 48.00 \\
Newsela ES 2-3 & es & 20,582 & 16,148 & 225,659 & 178,117 & 19.35 & 14.72 & 5.31 & 5.21 & 48.00 & 49.79 \\
Newsela ES 3-4 & es & 16,148 & 11,695 & 178,117 & 122,064 & 14.72 & 11.42 & 5.21 & 5.13 & 49.79 & 43.98 \\
Simplext & es & 8,071 & 3,191 & 38,731 & 25,409 & 34.96 & 14.59 & 5.47 & 5.34 & 5.74 & 9.03 \\
CBST Intuitive & eu & 1,697 & 1,586 & 4,575 & 4,447 & 19.98 & 14.53 & \textbf{6.34} & \textbf{6.43} & 76.33 & 102.00 \\
CBST Structural & eu & 1,697 & 1,654 & \textbf{4,575} & \textbf{4,793} & 19.98 & 16.82 & 6.34 & 6.29 & 76.33 & 95.00 \\
Alector & fr & 5,728 & 5,024 & 28,283 & 26,179 & 22.99 & 21.96 & 4.78 & 4.73 & 15.57 & 15.09 \\
CLEAR & fr & 11,743 & 11,205 & 119,465 & 118,212 & 25.99 & 25.72 & \textbf{5.72} & \textbf{5.73} & --- & --- \\
WikiLargeFR & fr & 205,933 & 173,827 & 8,763,745 & 6,384,020 & 28.54 & 20.70 & 5.03 & 4.94 & --- & --- \\
AdminIT & it & 3,420 & 3,394 & 29,581 & 28,784 & 38.07 & 37.72 & 6.08 & 5.90 & --- & --- \\
PaCCSS-IT & it & 10,478 & 9,853 & 580,389 & 519,211 & 9.21 & 8.24 & \textbf{4.75} & \textbf{4.79} & --- & --- \\
SimpitikiWiki & it & 9,188 & 9,175 & 41,899 & 41,375 & 72.87 & 71.96 & 5.60 & 5.60 & --- & --- \\
Teacher & it & 1,485 & 1,061 & 4,225 & 3,367 & 20.71 & 17.27 & 4.89 & 4.76 & 11.33 & 10.83 \\
Terence & it & 3,681 & 3,219 & 19,455 & 18,881 & 18.80 & 17.81 & 5.13 & 5.04 & 32.34 & 33.12 \\
Easy Japanese & ja & 10,331 & 3,401 & \textbf{489,302} & \textbf{517,651} & \textbf{9.79} & \textbf{10.35} & 1.51 & 1.49 & --- & --- \\
Easy Japanese Ext & ja & 18,888 & 5,305 & \textbf{433,341} & \textbf{503,035} & \textbf{12.38} & \textbf{14.37} & 1.55 & 1.49 & --- & --- \\
PorSimples Natural & pt-br & 9,983 & 9,527 & \textbf{64,610} & \textbf{65,174} & 20.97 & 13.52 & \textbf{5.39} & \textbf{5.53} & 20.01 & 31.31 \\
PorSimples Strong & pt-br & \textbf{9,527} & \textbf{9,601} & \textbf{65,174} & \textbf{65,552} & 13.52 & 12.25 & \textbf{5.53} & \textbf{5.57} & 31.31 & 34.76 \\
RSSE Corpus & ru & \textbf{16,467} & \textbf{24,307} & 138,319 & 95,067 & 20.33 & 13.97 & 6.73 & 6.54 & --- & --- \\
RuAdapt Ency A-B & ru & 4,609 & 3,842 & 11,085 & 9,804 & 12.58 & 9.96 & 6.08 & 5.84 & 14.44 & 16.13 \\
RuAdapt Ency A-C & ru & 4,927 & 3,844 & 11,931 & 9,809 & 13.59 & 9.96 & 6.05 & 5.84 & 14.16 & 15.89 \\
RuAdapt Ency B-C & ru & 27,268 & 26,200 & 113,817 & 110,463 & 14.28 & 13.37 & 6.10 & 6.08 & 29.96 & 31.06 \\
RuAdapt Fairytales & ru & 1,688 & 1,512 & 4,391 & 4,289 & 14.16 & 10.62 & \textbf{5.08} & \textbf{5.32} & 34.44 & 44.89 \\
RuAdapt Literature & ru & 55,321 & 42,655 & 368,499 & 327,228 & 15.26 & 11.58 & 5.14 & 5.08 & 168.90 & 197.62 \\
RuWikiLarge & ru & 331,063 & 275,644 & 5,760,207 & 4,540,009 & 20.70 & 15.68 & 5.95 & 5.79 & --- & --- \\
SloTS & sl & 5,871 & 2,723 & 21,804 & 10,646 & 18.46 & 8.27 & 4.72 & 4.44 & --- & --- \\
SimplifyUR & ur & \textbf{1,469} & \textbf{1,475} & 6,580 & 6,561 & 8.94 & 8.91 & 4.25 & 4.22 & --- & --- \\ \bottomrule
\end{tabular}}
\caption{\label{table:basic-stats}
Basic statistics about all of the corpora we analyzed.  Typically the vocab size, token count, average tokens per sentence, and average characters per token all decrease from original to simplified texts.  Outliers of this trend are highlighted in bold.
}
\end{table*}

\subsection{Document-level Compression}
In order to measure the editing levels on a document scale, we investigated document-level compression. Document-level compression is the ratio of the number of characters in the simple document to the number of characters in the original document \citep{xu-etal-2015-problems}.  A low document compression ratio indicates a lot of deletions between the original and simplified text, while a high document compression ratio suggests lengthier operations like sentence splitting or rephrasing.

\begin{figure*}[t!]
    \centering
    \includegraphics[width=\linewidth]{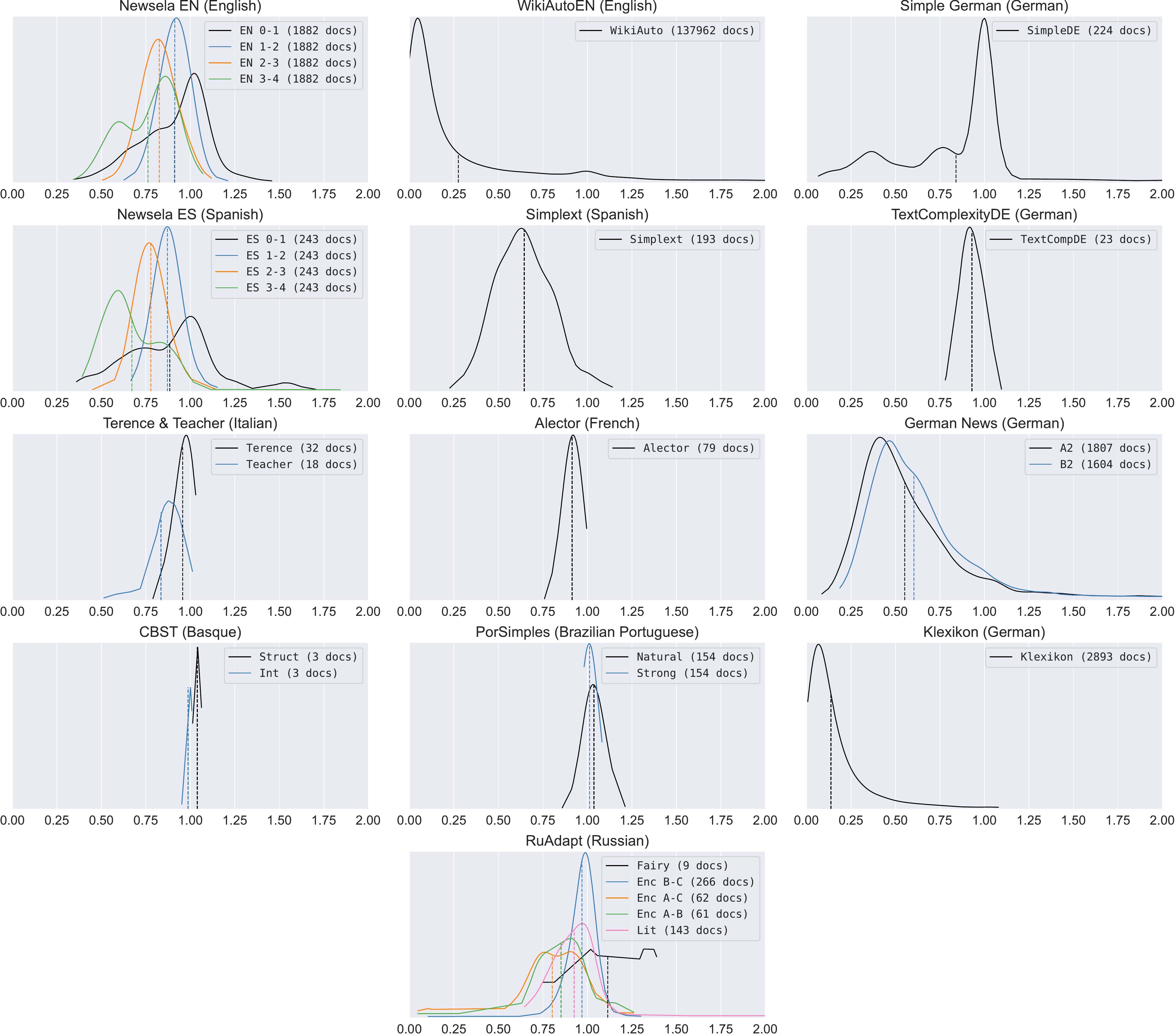}
    \caption{Distribution of document-level compression ratio for document-aligned corpora, smoothed by Gaussian kernel density estimation.  Means are marked by dashed lines.}
    \label{fig:compress_ratio}
\end{figure*}

We report the compression ratios of all document-aligned corpora in Figure \ref{fig:compress_ratio}.  Most of the compression ratios are approximately normally distributed.  For many of the corpora, the compression ratio is centered around one, meaning the original and simplified documents are about the same length.  This closely matches the low edit ratios that many of the corpora have (see section \ref{subsec:edit-distances}).  A few of the corpora (Simplext, Newsela, and German News) have lower means instead suggesting more significant document-level edits, such as deletion of entire sentences.

\subsection{Sentence-level Edit Operations}
Edit operations describe the types of simplifications that were performed to transform from an original sentence to a simplified sentence.  These can only be computed for corpora that are sentence aligned.  There are 6 edit operations we tracked from the alignments.  Each operation corresponds to a mapping ($x$:$y$) of $x$ original sentences to $y$ simple sentences. The operations are deletion (1:0), split (1:n), same (1:1), change (1:1), merge (n:1), and insert (0:1).  To determine the difference between ``same'' and ``change'', the Levenshtein distance \citep{Levenshtein1965BinaryCC} was measured between the original and simplified sentence.  This distance was divided by the length of the longer sentence.  If the difference was greater than 5\% then the sentences were marked as changed, otherwise, they were considered the same.  Levenshtein distance was calculated using the \textit{fuzzywuzzy}\footnote{\url{https://github.com/seatgeek/thefuzz}} library.

Table \ref{table:edit-ops} shows the distribution of edit operations.  Sentence-level edit operations were reported for both document-aligned corpora as well as sentence-aligned corpora that used sentence splitting (1:n mapping).  The most common edit operation across corpora was changing the original sentence, followed by keeping the same sentence, then splitting, deleting, and merging. Interestingly, Spanish corpora, like English ones \cite{xu-etal-2015-problems,jiang-etal-2020-neural}, had more deletion operations than most of the other languages.  For the sentence-only aligned corpora, this was because original sentences without simplifications were not included, but this was true even amongst document-aligned corpora.

\begin{table}[t!]
\small
\setlength{\tabcolsep}{2pt}
\renewcommand{\arraystretch}{0.5}
\centering
\resizebox{0.475\textwidth}{!}{%
\begin{tabular}{@{}l|ccccc|c@{}}
\toprule
\multicolumn{1}{c|}{\textbf{Corpus}} & \textbf{\begin{tabular}[c]{@{}c@{}}Deleted\\ 1:0 (\%)\end{tabular}} & \textbf{\begin{tabular}[c]{@{}c@{}}Split\\ 1:n (\%)\end{tabular}} & \textbf{\begin{tabular}[c]{@{}c@{}}Same\\ 1:1 (\%)\end{tabular}} & \textbf{\begin{tabular}[c]{@{}c@{}}Changed\\ 1:1 (\%)\end{tabular}} & \textbf{\begin{tabular}[c]{@{}c@{}}Merged\\ n:1 (\%)\end{tabular}} & \textbf{\begin{tabular}[c]{@{}c@{}}Inserted\\ 0:1 (\%)\end{tabular}} \\ \midrule
\textbf{English} & \multicolumn{1}{l}{} & \multicolumn{1}{l}{} & \multicolumn{1}{l}{} & \multicolumn{1}{l}{} & \multicolumn{1}{l|}{} & \multicolumn{1}{l}{} \\
Newsela EN 0-1 & 22.7 & 12.6 & 39.5 & 25.2 & 0.0 & 12.2 \\
Newsela EN 1-2 & 17.4 & 13.4 & 36.4 & 32.8 & 0.0 & 10.7 \\
Newsela EN 2-3 & 27.2 & 11.8 & 23.5 & 37.5 & 0.0 & 15.9 \\
Newsela EN 3-4 & 33.0 & 10.4 & 21.2 & 35.4 & 0.0 & 17.8 \\
WikiAuto & 94.2 & 0.9 & 1.2 & 3.7 & 0.0 & 44.8 \\ \midrule
\textbf{Russian} &  &  &  &  &  &  \\
RuAdapt Fairytales & 0.0 & 22.6 & 4.2 & 73.2 & 0.0 & 0.0 \\
RuAdapt Ency B-C & 0.0 & 3.5 & 79.3 & 17.2 & 0.0 & 0.0 \\
RuAdapt Ency A-C & 0.0 & 10.8 & 31.8 & 57.4 & 0.0 & 0.0 \\
RuAdapt Ency A-B & 0.0 & 10.7 & 35.4 & 53.9 & 0.0 & 0.0 \\
RuAdapt Literature & 0.0 & 11.8 & 36.6 & 51.6 & 0.0 & 0.0 \\ \midrule
\textbf{Italian} &  &  &  &  &  &  \\
Terence & 0.7 & 4.1 & 35.4 & 57.0 & 2.9 & 0.4 \\
Teacher & 6.9 & 9.3 & 8.3 & 59.3 & 16.2 & 1.5 \\ \midrule
\textbf{Spanish} &  &  &  &  &  &  \\
Newsela ES 0-1 & 29.1 & 20.0 & 19.1 & 31.6 & 0.2 & 0.7 \\
Newsela ES 1-2 & 18.3 & 19.8 & 24.2 & 37.7 & 0.1 & 0.5 \\
Newsela ES 2-3 & 27.5 & 22.9 & 13.3 & 36.0 & 0.3 & 0.4 \\
Newsela ES 3-4 & 38.5 & 19.1 & 11.2 & 31.1 & 0.2 & 0.3 \\
Simplext & 16.2 & 32.2 & 3.5 & 47.4 & 0.7 & 19.3 \\ \midrule
\textbf{German} &  &  &  &  &  &  \\
TextComplexityDE & 0.0 & 0.0 & 0.4 & 99.6 & 0.0 & 0.0 \\
German News A2 & 0.0 & 22.5 & 0.8 & 37.6 & 39.1 & 0.0 \\
German News B2 & 0.0 & 23.0 & 1.4 & 33.2 & 42.4 & 0.0 \\ \midrule
\textbf{Brazilian Portuguese} &  &  &  &  &  &  \\
PorSimples Natural & 0.6 & 38.7 & 22.1 & 38.3 & 0.3 & 0.8 \\
PorSimples Strong & 0.2 & 9.9 & 73.7 & 16.0 & 0.1 & 0.1 \\ \midrule
\textbf{Basque} &  &  &  &  &  &  \\
Structural CBST & 0.0 & 22.3 & 24.0 & 51.5 & 2.2 & 0.0 \\
Intuitive CBST & 0.0 & 25.8 & 27.1 & 45.9 & 1.3 & 0.0 \\ \bottomrule
\end{tabular}
}
\caption{\label{table:edit-ops}
Edit operations for all document-level corpora with sentence alignment.  All operations are reported as a percentage of original sentences besides ``inserted'' which is reported as a percentage of simplified sentences.
}
\end{table}

\begin{figure}[t!]
    \centering
    \includegraphics[width=\linewidth]{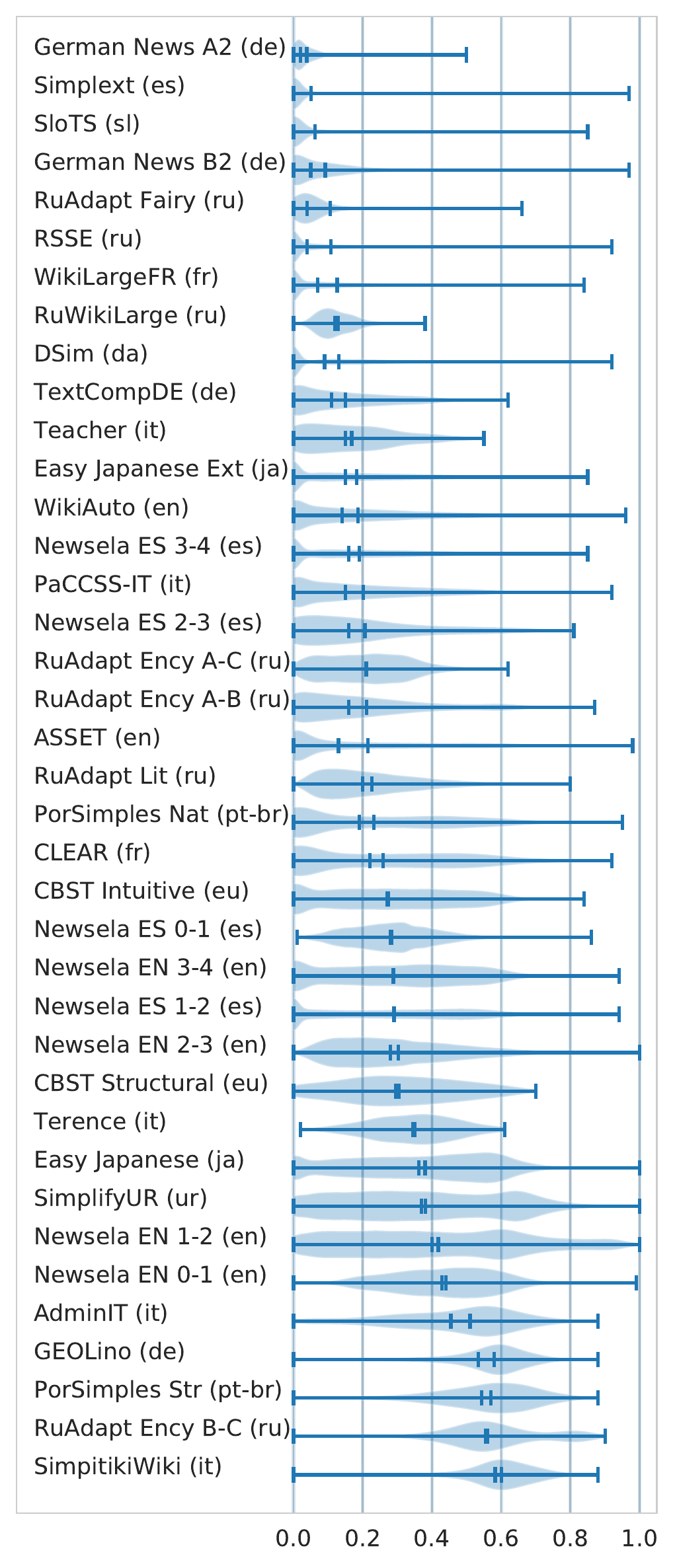}
    \caption{Violin Plots showing the minimum, maximum, mean, and median values for edit distances for all of the sentences in each corpus.  Distributions estimated using Gaussian kernel density estimation.}
    \label{fig:edit_distance}
\end{figure}

\subsection{Character-level Edit Distances}
\label{subsec:edit-distances}
We analyzed the edit distance distribution \citep{vasquez-rodriguez-etal-2021-investigating} on all corpora with sentence-level alignments to understand the strength of the edits at a sentence scale.  Low edit distances indicate smaller simplifications while high edit distances indicate big changes. We again used character-based Levenshtein distance to measure edit distance.  The Levenshtein distance was divided by the length of the longer sentence to obtain a ratio from zero to one.  Zero meant the sentence wasn't edited at all, while one meant the sentence was completely different.

The edit distance ratios can be found in Figure \ref{fig:edit_distance}.  For about half of the corpora, the mean edit distance fell below 20\%.  For the other half mean edit distance ratios ranged from 0.2 to 0.6.  Corpora with an approximately normal distribution and higher variance demonstrate a wide variety of both minor and major sentence edits.  Corpora with low means and a high concentration of low edit ratios primarily consist of slight modifications.

\section{Experimental Details}
\label{sec:experimental-details}

\textbf{Fine-tuning} For fine-tuning experiments we used the mT5 {\small Base} \citep{xue-etal-2021-mt5} architecture (580M Parameters).  We used the sentence piece tokenizer \citep{kudo-richardson-2018-sentencepiece} and limited inputs/targets to a length of 128 tokens.  We used a learning rate of \textit{5e-5} and the AdamW Optimizer \cite{loshchilov2018decoupled}.  Decoding was done using beam search with 4 beams.  The train batch size was set to 8.  We train for 5 epochs.  For any dataset without a development set, we removed 10\% of the training set up to 1,000 sentences to create our own dev set.  The training was performed on three NVIDIA A40 GPUs.

We also perform preprocessing on the training data inputs.  We compute sentence BLEU scores between the original and reference simplifications.  For any sentence pairs with an S-BLEU score below 10 or above 70, we remove it from the training set \citep{maddela-etal-2021-controllable}.  This helps reduce both misaligned and identical pairs.  Following \citet{martin-etal-2020-controllable} we also add control tokens to the input sentences.  We include a character length compression token \texttt{<NC\_[\#]>}, a Levenshtein similarity token \texttt{<LS\_[\#]>}, a dependency tree depth ratio token \texttt{<DR\_[\#]>}, and a word frequency rank token \texttt{<WR\_[\#]>}.  For each token, we compute the respective measure on both the original and simple sentences and include the ratio between [0.05, 2] in increments of 0.05.  For example, a prefix might have the form: "\texttt{<NC\_0.9>  <LS\_0.8>  <DR\_0.9>  <WR\_1.05>}".  More details on computing these tokens can be found in the original paper \citep{martin-etal-2020-controllable}.  For our dev set grid search to find the optimal token setting we perform a 3x3x3x3 search around the average values from the training set. \\

\noindent \textbf{Fewshot} For fewshot experiments we used BLOOM \citep{scao2022bloom} (176B Parameters).  We used the HuggingFace inference API\footnote{\url{https://huggingface.co/inference-api}} to prompt BLOOM with sampling.  We used a temperature of 1.0 and a repetition penalty of 0.0.  We formatted prompts to BLOOM as:

\begin{verbatim}
Original: "[EXAMPLE 1 ORIGINAL]"
Simple: "[EXAMPLE 1 SIMPLIFICATION]"
...
Original: "[EXAMPLE N ORIGINAL]"
Simple: "[EXAMPLE N ORIGINAL]"

Original: "[TEST ORIGINAL]"
Simple: "
\end{verbatim}
The prefixes "Original" and "Simple" were always in English.   Outputs were appended to the prompt and repeated back to the model until the output contained an end quotation followed by a new "\texttt{Original:}".  This was to prevent half-completed simplifications. \\ 

\noindent \textbf{Manual Evaluation} Table \ref{table:manual-eval-instructions} shows the key provided to the human annotators in each language.  Annotators were volunteers with fluency in the target language for annotation.  We randomly sampled 20 sentences from the test set of each of the four datasets and we used these 20 sentences to compare across all models.  For the "Reference" baseline if any sentence had more than one possible reference simplification we randomly sampled a reference.  For any of the system outputs, if the sentence was completely nonsense annotators were instructed to rate the sentence with a score of 1 on all aspects.

All annotators were volunteers that were informed that we were "measuring the quality of various machine learning models that have been trained/prompted to simplify text".  All of the volunteers were told they were assessing sentences to be used in evaluation for a research project. For any student employees that volunteered their time for  evaluation, they were paid at their normal hourly rate of \$18 per hour.  Other colleagues that volunteered their time did so on a strictly voluntary basis.

\begin{table*}
\small
\setlength{\tabcolsep}{2pt}
\renewcommand{\arraystretch}{0.5}
\centering
\resizebox{0.99\textwidth}{!}{%
\begin{tabular}{@{}cc|ccc|ccccc|c@{}}
\toprule
\multicolumn{2}{r|}{Lang $\rightarrow$} & \multicolumn{3}{c|}{en} & \multicolumn{5}{c|}{ru} & da \\ \midrule
Approach & \multicolumn{1}{l|}{\# shots} & ASSET & Newsela EN & WikiAuto EN & RuWikiLarge & RSSE & RuAdapt Ency & RuAdapt Fairy & RuAdapt Lit & DSim \\ \midrule
Fine-Tuned & NA & 35.98 & 38.60 & 42.46 & 32.01 & 31.66 & 26.42 & 34.79 & 41.75 & 31.40 \\ \midrule
Zero Shot & 0 & 35.52 & 33.22 & 34.73 & 31.47 & 20.09 & 33.20 & 12.74 & 30.95 & 35.84 \\ \midrule
 & 1 & 36.21 & 35.29 & 40.24 & 35.95 & 30.18 & 40.21 & 33.49 & 37.05 & 38.27 \\
 & 2 & 36.37 & 37.70 & 40.97 & 36.93 & 29.75 & 42.22 & 35.51 & 37.61 & 38.84 \\
 & 3 & 36.53 & 37.71 & 41.66 & 37.22 & 29.79 & 40.76 & 37.07 & 39.06 & 38.09 \\
 & 5 & 36.79 & 38.82 & 42.25 & 36.84 & 31.08 & 41.01 & 38.43 & 39.89 & 37.57 \\
 & 10 & 36.19 & 38.65 & 42.66 & 37.59 & 31.33 & 39.45 & 40.93 & 40.44 & 31.71 \\
\multirow{-9}{*}{\begin{tabular}[c]{@{}c@{}}Semantic\\ Similarity\end{tabular}} & 20 & 36.80 & 39.26 & 42.83 & 37.71 & 31.22 & 39.54 & 38.95 & 40.32 & 29.88 \\ \midrule
 & 1 & 35.21 & 34.02 & 35.16 & 33.61 & 28.23 & 33.45 & 20.89 & 32.26 & 34.96 \\
 & 2 & 35.77 & 34.87 & 36.40 & 34.20 & 28.44 & 34.07 & 24.24 & 33.14 & 35.29 \\
 & 3 & 35.37 & 34.31 & 36.21 & 34.19 & 29.40 & 34.24 & 22.84 & 33.41 & 35.39 \\
 & 5 & 35.89 & 34.39 & 36.00 & 32.89 & 29.30 & 33.17 & 25.91 & 33.45 & 34.81 \\
 & 10 & 36.10 & 35.35 & 36.86 & 34.63 & 28.69 & 33.19 & 29.16 & 33.92 & 30.15 \\
\multirow{-9}{*}{\begin{tabular}[c]{@{}c@{}}Random\\ Sampling\end{tabular}} & 20 & 36.21 & 34.73 & 37.14 & 34.63 & 29.60 & 33.71 & 31.83 & 33.89 & 27.98 \\ \midrule
\multicolumn{2}{r|}{Lang $\rightarrow$} & \multicolumn{3}{c|}{de} & \multicolumn{5}{c|}{it} & pt-br \\ \midrule
Approach & \multicolumn{1}{l|}{\# shots} & German News & TextCompDE & GEOLino & PaCCSS-IT & Terence & AdminIT & Simpitiki & Teacher & PorSimples \\ \midrule
Fine-Tuned & NA & 36.04 & 30.26 & 26.44 & 57.30 & 23.92 & 23.42 & 4.11 & 29.84 & 31.54 \\ \midrule
Zero Shot & 0 & 32.48 & 32.26 & 29.59 & 35.42 & 35.91 & 32.43 & 18.43 & 28.75 & 35.38 \\ \midrule
 & 1 & 36.19 & 37.63 & 38.16 & 51.42 & 34.95 & 37.06 & 27.73 & 33.97 & 36.72 \\
 & 2 & 36.68 & 38.60 & 39.65 & 49.15 & 37.25 & 36.69 & 26.42 & 34.14 & 37.46 \\
 & 3 & 36.78 & 41.03 & 39.44 & 48.00 & 34.95 & 35.67 & 27.06 & 29.41 & 38.85 \\
 & 5 & 37.79 & 38.81 & 39.5 & 45.48 & 35.94 & 38.16 & 26.94 & 39.10 & 38.96 \\
 & 10 & 37.69 & 38.93 & 39.7 & 37.31 & 35.39 & 35.21 & 27.20 & 32.62 & 41.34 \\
\multirow{-9}{*}{\begin{tabular}[c]{@{}c@{}}Semantic\\ Similarity\end{tabular}} & 20 & 36.76 & 38.93 & 39.44 & 33.45 & 35.17 & 35.21 & 27.73 & 33.46 & 39.94 \\ \midrule
 & 1 & 32.58 & 34.94 & 35.89 & 38.84 & 33.60 & 33.31 & 21.96 & 33.94 & {\color[HTML]{1D1C1D} 37.68} \\
 & 2 & 34.09 & 35.37 & 36.11 & 39.11 & 34.15 & 34.77 & 23.79 & 25.01 & {\color[HTML]{1D1C1D} 36.82} \\
 & 3 & 34.92 & 34.13 & 35.22 & 38.51 & 33.96 & 35.98 & 25.22 & 31.41 & {\color[HTML]{1D1C1D} 36.57} \\
 & 5 & 34.71 & 36.68 & 34.5 & 37.41 & 32.01 & 34.24 & 25.01 & 32.30 & {\color[HTML]{1D1C1D} 36.53} \\
 & 10 & 35.58 & 38.07 & 35.42 & 35.01 & 31.60 & 35.67 & 25.04 & 30.82 & {\color[HTML]{1D1C1D} 35.93} \\
\multirow{-9}{*}{\begin{tabular}[c]{@{}c@{}}Random\\ Sampling\end{tabular}} & 20 & 35.53 & 38.07 & 34.62 & 30.29 & 34.38 & 35.67 & 25.04 & 34.39 & {\color[HTML]{1D1C1D} 35.31} \\ \midrule
\multicolumn{2}{r|}{Lang $\rightarrow$} & \multicolumn{2}{c|}{fr} & sl & \multicolumn{2}{c|}{ja} & \multicolumn{2}{c|}{es} & ur & eu \\ \midrule
Approach & \multicolumn{1}{l|}{\# shots} & WikiLargeFR & \multicolumn{1}{c|}{CLEAR} & SloTS & Easy JA & \multicolumn{1}{c|}{Easy JA Ext} & NewselaES & \multicolumn{1}{c|}{Simplext} & SimplifyUR & CBST \\ \midrule
Fine-Tuned & NA & 35.20 & \multicolumn{1}{c|}{34.86} & 36.56 & 67.36 & \multicolumn{1}{c|}{43.15} & 29.89 & \multicolumn{1}{c|}{35.62} & 33.18 & 24.26 \\ \midrule
Zero Shot & 0 & 35.71 & \multicolumn{1}{c|}{35.75} & 27.37 & 41.71 & \multicolumn{1}{c|}{30.53} & 34.15 & \multicolumn{1}{c|}{25.36} & 28.61 & 34.43 \\ \midrule
 & 1 & 36.29 & \multicolumn{1}{c|}{38.06} & 30.75 & 48.71 & \multicolumn{1}{c|}{46.08} & 37.07 & \multicolumn{1}{c|}{32.50} & 39.12 & 35.09 \\
 & 2 & 35.22 & \multicolumn{1}{c|}{39.03} & 37.38 & 54.89 & \multicolumn{1}{c|}{49.39} & 36.90 & \multicolumn{1}{c|}{38.05} & 50.69 & 38.89 \\
 & 3 & 36.40 & \multicolumn{1}{c|}{40.16} & 37.20 & 55.38 & \multicolumn{1}{c|}{47.01} & 38.18 & \multicolumn{1}{c|}{39.28} & 51.23 & 38.84 \\
 & 5 & 36.75 & \multicolumn{1}{c|}{39.34} & 37.55 & 57.29 & \multicolumn{1}{c|}{49.30} & 38.12 & \multicolumn{1}{c|}{40.26} & 53.60 & 39.50 \\
 & 10 & 36.33 & \multicolumn{1}{c|}{39.21} & 36.91 & 58.67 & \multicolumn{1}{c|}{47.50} & 38.42 & \multicolumn{1}{c|}{39.90} & 55.89 & 37.30 \\
\multirow{-9}{*}{\begin{tabular}[c]{@{}c@{}}Semantic\\ Similarity\end{tabular}} & 20 & 37.72 & \multicolumn{1}{c|}{38.45} & 37.24 & 59.42 & \multicolumn{1}{c|}{46.55} & 38.42 & \multicolumn{1}{c|}{39.75} & 56.36 & 38.31 \\ \midrule
 & 1 & 35.64 & \multicolumn{1}{c|}{36.12} & 27.43 & 38.35 & \multicolumn{1}{c|}{40.70} & {\color[HTML]{1D1C1D} 34.25} & \multicolumn{1}{c|}{{\color[HTML]{1D1C1D} 29.57}} & 40.38 & 36.40 \\
 & 2 & 36.67 & \multicolumn{1}{c|}{34.64} & 31.83 & 37.85 & \multicolumn{1}{c|}{41.38} & {\color[HTML]{1D1C1D} 34.55} & \multicolumn{1}{c|}{{\color[HTML]{1D1C1D} 33.51}} & 49.00 & 36.00 \\
 & 3 & 36.07 & \multicolumn{1}{c|}{36.92} & 33.95 & 35.90 & \multicolumn{1}{c|}{40.04} & {\color[HTML]{1D1C1D} 34.16} & \multicolumn{1}{c|}{{\color[HTML]{1D1C1D} 35.56}} & 49.28 & 36.10 \\
 & 5 & 36.17 & \multicolumn{1}{c|}{35.54} & 33.60 & 35.92 & \multicolumn{1}{c|}{42.55} & {\color[HTML]{1D1C1D} 33.64} & \multicolumn{1}{c|}{{\color[HTML]{1D1C1D} 36.81}} & 50.34 & 37.39 \\
 & 10 & 36.70 & \multicolumn{1}{c|}{35.80} & 34.18 & 39.34 & \multicolumn{1}{c|}{42.50} & {\color[HTML]{1D1C1D} 34.27} & \multicolumn{1}{c|}{{\color[HTML]{1D1C1D} 37.40}} & 52.93 & 36.83 \\
\multirow{-9}{*}{\begin{tabular}[c]{@{}c@{}}Random\\ Sampling\end{tabular}} & 20 & 35.80 & \multicolumn{1}{c|}{36.13} & 35.30 & 37.89 & \multicolumn{1}{c|}{42.11} & {\color[HTML]{1D1C1D} 33.91} & \multicolumn{1}{c|}{{\color[HTML]{1D1C1D} 39.27}} & 50.02 & 38.86 \\ \bottomrule
\end{tabular}
}
\caption{\label{table:fewshot}
SARI Scores for BLOOM Fewshot Experiments
}
\end{table*}

\begin{table*}
\small
\setlength{\tabcolsep}{2pt}
\renewcommand{\arraystretch}{1.0}
\centering
\resizebox{0.8\textwidth}{!}{%
\begin{tabular}{@{}cl@{}}
\toprule
\multicolumn{2}{c}{\textbf{Adequacy (is the meaning preserved?)}} \\ \midrule
\;\;\;1:\;\;\; & The subject of the sentence has changed entirely and is entirely unrelated \\
2: & The meaning has been seriously altered (negated or changed) \\
3: & \textbf{Two or more important pieces} of information have been added or removed \\
4: & Meaning is similar but \textbf{one} piece of information has been added or removed \\
5: & Meaning is preserved aside from minor unimportant information \\ \midrule
\multicolumn{2}{c}{\textbf{Fluency (is the simplification eloquent/grammatical?)}} \\ \midrule
1: & The simplification is completely unreadable \\
2: & The simplification suffers from many serious grammar issues (nearly unreadable) \\
3: & The simplification has \textbf{two or more} grammatical mistakes \\
4: & The simplification has a minor grammatical issue or is written strangely in one place \\
5: & The simplification is perfectly eloquent as if written by a human \\ \midrule
\multicolumn{2}{c}{\textbf{Simplicity (is the simplification actually simpler?)}} \\ \midrule
1: & The simplification is actually harder to understand (ex. more complex terms used) \\
2: & The simplification is about the same difficulty as the original \\
3: & The simplification is mildly simpler, but this simplification does not help readability \\
4: & The simplification is actually simpler \\
5: & The simplification is vastly simpler and could help someone better understand \\ \bottomrule
\end{tabular}
}
\caption{\label{table:manual-eval-instructions}
Manual evaluation key provided to annotators
}
\end{table*}

\clearpage
\onecolumn
{\small
\centering

\begin{longtable}{@{}lll@{}}
\toprule
 & \multicolumn{1}{c}{Sentence} & \multicolumn{1}{c}{Translated} \\* \midrule
\endfirsthead
\multicolumn{3}{c}%
{{\bfseries Table \thetable\ continued from previous page}} \\
\toprule
 & \multicolumn{1}{c}{Sentence} & \multicolumn{1}{c}{Translated} \\* \midrule
\endhead
\multicolumn{3}{c}{\textbf{AdminIT (Italian)}} \\* \midrule
\textbf{Original} & \multicolumn{1}{l|}{\begin{tabular}[c]{@{}l@{}}E’ presente anche il personale esecutivo che \\ provvede allo sporzionamento delle portate.\end{tabular}} & \begin{tabular}[c]{@{}l@{}}There is also the executive staff who arrange \\ portioning of the courses.\end{tabular} \\* \midrule
\textbf{Simple} & \multicolumn{1}{l|}{\begin{tabular}[c]{@{}l@{}}È presente anche il personale che divide le portate \\ in porzioni.\end{tabular}} & \begin{tabular}[c]{@{}l@{}}There is also the staff who divide the courses into \\ portions.\end{tabular} \\* \midrule
\multicolumn{3}{c}{\textbf{ASSET (English)}} \\* \midrule
\textbf{Original} & \multicolumn{1}{l|}{\begin{tabular}[c]{@{}l@{}}The Apostolic Tradition, attributed to the theologian \\ Hippolytus, attests the singing of Hallel psalms with \\ Alleluia as the refrain in early Christian agape feasts.\end{tabular}} &  \\* \midrule
\textbf{Simple} & \multicolumn{1}{l|}{\begin{tabular}[c]{@{}l@{}}The Apostolic Tradition was created by the religion \\ expert Hippolytus. It shows the singing of Hallel \\ psalms with Alleluia as the refrain in early Christian \\ feasts.\end{tabular}} &  \\* \midrule
\multicolumn{3}{c}{\textbf{CBST (Basque)}} \\* \midrule
\textbf{Original} & \multicolumn{1}{l|}{\begin{tabular}[c]{@{}l@{}}Horrekin batera, gure planeta eta eguzki-sistema \\ gainerakoekin alderatu nahi dituzte, eta ikusi nahi \\ dute ea horrelakoak fenomeno bakanak diren edo \\ oso arruntak diren unibertsoan.\end{tabular}} & \begin{tabular}[c]{@{}l@{}}At the same time, they want to compare our planet \\ and the solar system with the rest, and they want \\ to see if such phenomena are rare or very common \\ in the universe.\end{tabular} \\* \midrule
\textbf{Simple} & \multicolumn{1}{l|}{\begin{tabular}[c]{@{}l@{}}Gainera, gure planeta eta eguzki-sistema \\ gainerakoekin alderatu nahi dituzte; \\ fenomeno horiek bakanak edo arruntak \\ dira unibertsoan? hori ikusi nahi dute.\end{tabular}} & \begin{tabular}[c]{@{}l@{}}They also want to compare our planet and the solar \\ system with the rest; Are these phenomena rare \\ or common in the universe? they want to see that.\end{tabular} \\* \midrule
\multicolumn{3}{c}{\textbf{CLEAR (French)}} \\* \midrule
\textbf{Original} & \multicolumn{1}{l|}{\begin{tabular}[c]{@{}l@{}}une étude concernant les entretiens motivationnels \\ suggérait que cette intervention était bénéfique \\ contre la consommation de cannabis\end{tabular}} & \begin{tabular}[c]{@{}l@{}}a study on motivational interviewing suggested that \\ this intervention was beneficial against cannabis use\end{tabular} \\* \midrule
\textbf{Simple} & \multicolumn{1}{l|}{\begin{tabular}[c]{@{}l@{}}l' une des deux études concernant des entretiens \\ motivationnels suggérait que cette intervention \\ était bénéfique sur la consommation de cannabis \\ signalée\end{tabular}} & \begin{tabular}[c]{@{}l@{}}one of two studies involving motivational \\ interviewing suggested that this intervention \\ was beneficial on reported cannabis use\end{tabular} \\* \midrule
\multicolumn{3}{c}{\textbf{DSim (Danish)}} \\* \midrule
\textbf{Original} & \multicolumn{1}{l|}{\begin{tabular}[c]{@{}l@{}}Stigende vandstand i floderne i det østlige Tjekkiet \\ forvandlede i aftes hundredvis af boliger i området \\ til dødsfælder .\end{tabular}} & \begin{tabular}[c]{@{}l@{}}Rising water levels in the rivers in the eastern Czech \\ Republic last night turned hundreds of homes in the \\ area into death traps.\end{tabular} \\* \midrule
\textbf{Simple} & \multicolumn{1}{l|}{I det østlige Tjekkiet stiger vandstanden i floderne .} & \begin{tabular}[c]{@{}l@{}}In the eastern Czech Republic, the water level in the \\ rivers is rising.\end{tabular} \\* \midrule
\multicolumn{3}{c}{\textbf{EasyJA (Japanese)}} \\* \midrule
\textbf{Original} & \multicolumn{1}{l|}{\begin{CJK}{UTF8}{min} 君が言ったことで、僕はびっくりした。 \end{CJK}} & What you said surprised me. \\* \midrule
\textbf{Simple} & \multicolumn{1}{l|}{\begin{CJK}{UTF8}{min} あなたが言ったことで、私は驚いた。\end{CJK}} & What you said surprised me. \\* \midrule
\multicolumn{3}{c}{\textbf{EasyJAExt (Japanese)}} \\* \midrule
\textbf{Original} & \multicolumn{1}{l|}{\begin{CJK}{UTF8}{min} 彼の不注意にはあきれてしまった。 \end{CJK}} & I was appalled at his carelessness. \\* \midrule
\textbf{Simple} & \multicolumn{1}{l|}{\begin{CJK}{UTF8}{min} 彼の不注意には言葉を失う。 \end{CJK}} & His carelessness leaves me speechless. \\* \midrule
\multicolumn{3}{c}{\textbf{GEOLino (German)}} \\* \midrule
\textbf{Original} & \multicolumn{1}{l|}{\begin{tabular}[c]{@{}l@{}}Denn sie sind zwar mutig, aber durchaus nicht \\ lebensmüde.\end{tabular}} & \begin{tabular}[c]{@{}l@{}}Because they are courageous, but by no means tired \\ of life.\end{tabular} \\* \midrule
\textbf{Simple} & \multicolumn{1}{l|}{Denn sie sind zwar mutig, aber nicht lebensmüde.} & Because they are courageous, but not tired of life. \\* \midrule
\multicolumn{3}{c}{\textbf{GermanNews (German)}} \\* \midrule
\textbf{Original} & \multicolumn{1}{l|}{\begin{tabular}[c]{@{}l@{}}Jedes Kalb erhält spätestens sieben Tage nach der \\ Geburt eine eindeutig identifizierbare \\ Lebensnummer, die in Form von Ohrmarken \\ beidseitig eingezogen wird.\end{tabular}} & \begin{tabular}[c]{@{}l@{}}Each calf receives a clearly identifiable life number \\ no later than seven days after birth, which is recorded \\ on both sides in the form of ear tags.\end{tabular} \\* \midrule
\textbf{Simple} & \multicolumn{1}{l|}{\begin{tabular}[c]{@{}l@{}}In Österreich bekommt jedes Kalb kurz nach der \\ Geburt eine Nummer\end{tabular}} & \begin{tabular}[c]{@{}l@{}}In Austria, every calf is given a number shortly \\ after birth.\end{tabular} \\* \midrule
\multicolumn{3}{c}{\textbf{NewselaEN (English)}} \\* \midrule
\textbf{Original} & \multicolumn{1}{l|}{\begin{tabular}[c]{@{}l@{}}Putting these parts into jet engines is just what the \\ advanced manufacturing industry has been waiting \\ for: evidence that shows that mainstream \\ manufacturers have figured out how to make the \\ materials and the process work.\end{tabular}} &  \\* \midrule
\textbf{Simple} & \multicolumn{1}{l|}{\begin{tabular}[c]{@{}l@{}}Putting these parts into jet engines shows that \\ companies have figured out how to make the \\ materials and the process work.\end{tabular}} &  \\* \midrule
\multicolumn{3}{c}{\textbf{NewselaES (Spanish)}} \\* \midrule
\textbf{Original} & \multicolumn{1}{l|}{\begin{tabular}[c]{@{}l@{}}Para el proyecto de Apple, Taylor-Young tomó fotos \\ de paisajes urbanos bajo la lluvia con su iPhone 6.\end{tabular}} & \begin{tabular}[c]{@{}l@{}}For the Apple project, Taylor-Young took photos of \\ cityscapes in the rain with her iPhone 6.\end{tabular} \\* \midrule
\textbf{Simple} & \multicolumn{1}{l|}{\begin{tabular}[c]{@{}l@{}}Para el proyecto de Apple, utilizó su iPhone 6 \\ para tomar fotografías de las calles lluviosas \\ de la ciudad.\end{tabular}} & \begin{tabular}[c]{@{}l@{}}For the Apple project, she used her iPhone 6 to take \\ pictures of the rainy streets of the city.\end{tabular} \\* \midrule
\multicolumn{3}{c}{\textbf{PaCCSS-IT (Italian)}} \\* \midrule
\textbf{Original} & \multicolumn{1}{l|}{\begin{tabular}[c]{@{}l@{}}Anche per questa si chiede l' immediata \\ eseguibilità : Chi è favorevole ?\end{tabular}} & \begin{tabular}[c]{@{}l@{}}For this too , immediate execution is requested : \\ Who is in favor ?\end{tabular} \\* \midrule
\textbf{Simple} & \multicolumn{1}{l|}{\begin{tabular}[c]{@{}l@{}}Chiedo l' immediata eseguibilità : Chi è \\ favorevole ?\end{tabular}} & I ask for immediate execution : Who is in favor ? \\* \midrule
\multicolumn{3}{c}{\textbf{PorSimples (Brazilian Portuguese)}} \\* \midrule
\textbf{Original} & \multicolumn{1}{l|}{\begin{tabular}[c]{@{}l@{}}No Eldorado do Sul poderá ser construído um \\ estádio provisório.\end{tabular}} & \begin{tabular}[c]{@{}l@{}}In Eldorado do Sul, a provisional stadium could \\ be built.\end{tabular} \\* \midrule
\textbf{Simple} & \multicolumn{1}{l|}{\begin{tabular}[c]{@{}l@{}}No Eldorado do Sul talvez construam um \\ estádio provisório.\end{tabular}} & \begin{tabular}[c]{@{}l@{}}In Eldorado do Sul, perhaps they will build a \\ temporary stadium.\end{tabular} \\* \midrule
\multicolumn{3}{c}{\textbf{RSSE (Russian)}} \\* \midrule
\textbf{Original} & \multicolumn{1}{l|}{\begin{tabular}[c]{@{}l@{}}{\fontencoding{T2A}\selectfont В природном очаге заражение обычно }\\ {\fontencoding{T2A}\selectfont происходит через укус блохи, ранее }\\ {\fontencoding{T2A}\selectfont питавшейся на больном грызуне.}\end{tabular}} & \begin{tabular}[c]{@{}l@{}}In a natural focus, infection usually occurs through \\ the bite of a flea that previously fed on a sick rodent.\end{tabular} \\* \midrule
\textbf{Simple} & \multicolumn{1}{l|}{\begin{tabular}[c]{@{}l@{}}{\fontencoding{T2A}\selectfont Блоха может заразить укусом, если }\\ {\fontencoding{T2A}\selectfont ранее она кусала больного грызуна.}\end{tabular}} & \begin{tabular}[c]{@{}l@{}}A flea can infect with a bite if it has previously \\ bitten a sick rodent.\end{tabular} \\* \midrule
\multicolumn{3}{c}{\textbf{RuAdaptEncy (Russian)}} \\* \midrule
\textbf{Original} & \multicolumn{1}{l|}{\begin{tabular}[c]{@{}l@{}}{\fontencoding{T2A}\selectfont Достоевский женился на стенографистке }\\ {\fontencoding{T2A}\selectfont Анне Григорьевне Сниткиной, которая }\\ {\fontencoding{T2A}\selectfont стала ему близким другом и помощником.}\end{tabular}} & \begin{tabular}[c]{@{}l@{}}Dostoevsky married the stenographer Anna \\ Grigorievna Snitkina, who became his close friend \\ and assistant.\end{tabular} \\* \midrule
\textbf{Simple} & \multicolumn{1}{l|}{\begin{tabular}[c]{@{}l@{}}{\fontencoding{T2A}\selectfont Достоевский женился на Анне }\\ {\fontencoding{T2A}\selectfont Григорьевне Сниткиной.}\end{tabular}} & Dostoevsky married Anna Grigoryevna Snitkina. \\* \midrule
\multicolumn{3}{c}{\textbf{RuAdaptFairytales (Russian)}} \\* \midrule
\textbf{Original} & \multicolumn{1}{l|}{\begin{tabular}[c]{@{}l@{}}{\fontencoding{T2A}\selectfont Пустил стрелу средний брат — полетела }\\ {\fontencoding{T2A}\selectfont стрела к богатому купцу во двор.}\end{tabular}} & \begin{tabular}[c]{@{}l@{}}The middle brother fired an arrow - an arrow flew \\ to the rich merchant in the yard.\end{tabular} \\* \midrule
\textbf{Simple} & \multicolumn{1}{l|}{\begin{tabular}[c]{@{}l@{}}{\fontencoding{T2A}\selectfont Стрела среднего брата прилетела на }\\ {\fontencoding{T2A}\selectfont богатый купеческий двор.}\end{tabular}} & \begin{tabular}[c]{@{}l@{}}The arrow of the middle brother flew to the rich \\ merchant's yard.\end{tabular} \\* \midrule
\multicolumn{3}{c}{\textbf{RuAdaptLit (Russian)}} \\* \midrule
\textbf{Original} & \multicolumn{1}{l|}{\begin{tabular}[c]{@{}l@{}}{\fontencoding{T2A}\selectfont Попала бы моя книжка в лапки какой-нибудь }\\ {\fontencoding{T2A}\selectfont девочке в зеленом платьице… Села бы она у }\\ {\fontencoding{T2A}\selectfont камина с моим сочинением, читала бы, }\\ {\fontencoding{T2A}\selectfont перелистывала бы и улыбалась.}\end{tabular}} & \begin{tabular}[c]{@{}l@{}}My book would fall into the paws of some girl in \\ a green dress ... She would sit by the fireplace \\ with my essay, read, leaf through and smile.\end{tabular} \\* \midrule
\textbf{Simple} & \multicolumn{1}{l|}{\begin{tabular}[c]{@{}l@{}}{\fontencoding{T2A}\selectfont И какая-нибудь девочка сидела бы у камина }\\ {\fontencoding{T2A}\selectfont с моей книжкой, читала бы и улыбалась.}\end{tabular}} & \begin{tabular}[c]{@{}l@{}}And some girl would sit by the fireplace with my \\ book, read and smile.\end{tabular} \\*

\midrule
\multicolumn{3}{c}{\textbf{RuWikiLarge (Russian)}} \\* \midrule
\textbf{Original} & \multicolumn{1}{l|}{\begin{tabular}[c]{@{}l@{}}{\fontencoding{T2A}\selectfont Он служил во французском флоте, а в 1889}\\ {\fontencoding{T2A}\selectfont  и 1890 годах служил в команде фрегата Iphig}\\ {\fontencoding{T2A}\selectfont nie и несколько лет провел в Кочинчине. }\end{tabular}} & \begin{tabular}[c]{@{}l@{}}He served in the French Navy and in 1889 and 1890 \\  was in command of the frigate Iphig nie and spent \\ several years in Cochinchina.\end{tabular} \\* \midrule
\textbf{Simple} & \multicolumn{1}{l|}{\begin{tabular}[c]{@{}l@{}}{\fontencoding{T2A}\selectfont Некоторое время он провел во французском}\\ {\fontencoding{T2A}\selectfont флоте. В 1889 и 1890 годах он служил в}\\ {\fontencoding{T2A}\selectfont  команде фрегата Iphig nie.}\end{tabular}} & \begin{tabular}[c]{@{}l@{}}He spent some time in the French Navy. In 1889 and \\1890 he served in command of the frigate Iphig nie.\end{tabular} \\*

\midrule
\multicolumn{3}{c}{\textbf{Simpitiki Wiki (Italian)}} \\* \midrule
\textbf{Original} & \multicolumn{1}{l|}{\begin{tabular}[c]{@{}l@{}}Mesero (Mésar nella variante locale del dialetto \\ milanese) è un comune di 3.716 abitanti della \\ provincia di Milano.\end{tabular}} & \begin{tabular}[c]{@{}l@{}}Mesero (Mésar in the local variant of the Milanese \\ dialect) is a town of 3,716 inhabitants in the \\ province of Milan.\end{tabular} \\* \midrule
\textbf{Simple} & \multicolumn{1}{l|}{\begin{tabular}[c]{@{}l@{}}Mesero (Mésar nel locale dialetto milanese) è un \\ comune di 3.716 abitanti della provincia di \\ Milano.\end{tabular}} & \begin{tabular}[c]{@{}l@{}}Mesero (Mésar in the local Milanese dialect) is a \\ town of 3,716 inhabitants in the province of Milan.\end{tabular} \\* \midrule
\multicolumn{3}{c}{\textbf{Simplext (Spanish)}} \\* \midrule
\textbf{Original} & \multicolumn{1}{l|}{\begin{tabular}[c]{@{}l@{}}Oxfam señaló que las bajas temperaturas de este \\ invierno han aumentado el número de infecciones \\ respiratorias , como la gripe y la neumonía , con \\ más de 200.000 casos notificados en la segunda \\ semana de este mes de enero , y grandes \\ extensiones de tierra de el sur de Pakistán \\ continúan bajo el agua contaminada .\end{tabular}} & \begin{tabular}[c]{@{}l@{}}Oxfam said this winter's low temperatures have \\ increased the number of respiratory infections, \\ including influenza and pneumonia, with more \\ than 200,000 cases reported in the second week \\ of January, across large swaths of southern \\ Pakistan. they continue under polluted water.\end{tabular} \\* \midrule
\textbf{Simple} & \multicolumn{1}{l|}{\begin{tabular}[c]{@{}l@{}}Debido a el frío de el invierno , las enfermedades \\ han aumentado entre las personas de Pakistán.\end{tabular}} & \begin{tabular}[c]{@{}l@{}}Due to the cold of winter, diseases have increased \\ among the people of Pakistan.\end{tabular} \\* \midrule
\multicolumn{3}{c}{\textbf{SimplifyUR (Urdu)}} \\* \midrule
\textbf{Original} & \multicolumn{1}{r|}{\<اسے بولنے میں دشواری ہو رہی تھی>} & He was having trouble speaking \\* \midrule
\textbf{Simple} & \multicolumn{1}{r|}{\<اسے بولنے میں مشکل ہو رہی تھی>} & He was having difficulty speaking \\* \midrule
\multicolumn{3}{c}{\textbf{SloTS (Slovene)}} \\* \midrule
\textbf{Original} & \multicolumn{1}{l|}{\begin{tabular}[c]{@{}l@{}}Komaj sta bila v stolpu, je Hubert priskočil in \\ kmetico udaril v obraz. Nato jo je še sunil v \\ trebuh s svojim težkim škornjem.\end{tabular}} & \begin{tabular}[c]{@{}l@{}}As soon as they were in the tower, Hubert jumped \\ up and punched the peasant in the face. Then \\ he pushed her in the stomach with his heavy boot.\end{tabular} \\* \midrule
\textbf{Simple} & \multicolumn{1}{l|}{\begin{tabular}[c]{@{}l@{}}Ko jo je Hubert pripeljal v stolp, jo je udaril v \\ obraz in brcnil v trebuh.\end{tabular}} & \begin{tabular}[c]{@{}l@{}}When Hubert brought her to the tower, he punched \\ her in the face and kicked her in the stomach.\end{tabular} \\* \midrule
\multicolumn{3}{c}{\textbf{Teacher (Italian)}} \\* \midrule
\textbf{Original} & \multicolumn{1}{l|}{\begin{tabular}[c]{@{}l@{}}Sebbene sia umido, credo che ad Amsterdam \\ non abbiamo mai costruito niente di più \\ comodo per chi ha bisogno di nascondersi.\end{tabular}} & \begin{tabular}[c]{@{}l@{}}Although it is humid, I believe that in Amsterdam \\ we have never built anything more comfortable \\ for those who need to hide.\end{tabular} \\* \midrule
\textbf{Simple} & \multicolumn{1}{l|}{E’ umido ma è comodo come nascondiglio.} & It's humid but it's comfortable as a hiding place. \\* \midrule
\multicolumn{3}{c}{\textbf{Terence (Italian)}} \\* \midrule
\textbf{Original} & \multicolumn{1}{l|}{\begin{tabular}[c]{@{}l@{}}Tutti si precipitarono verso il tendone e si \\ ammassarono dentro per trovare riparo, \\ perché nessuno si voleva infradiciare.\end{tabular}} & \begin{tabular}[c]{@{}l@{}}Everyone rushed to the tent and crowded inside \\ for shelter, because no one wanted to get soaked.\end{tabular} \\* \midrule
\textbf{Simple} & \multicolumn{1}{l|}{\begin{tabular}[c]{@{}l@{}}Tutti si misero a correre verso la tenda, e ben \\ presto la tenda fu piena di gente, perché \\ nessuno si voleva bagnare.\end{tabular}} & \begin{tabular}[c]{@{}l@{}}Everyone ran towards the tent, and soon the tent \\ was full of people, because nobody wanted to \\ get wet.\end{tabular} \\* \midrule
\multicolumn{3}{c}{\textbf{TextComplexityDE (German)}} \\* \midrule
\textbf{Original} & \multicolumn{1}{l|}{\begin{tabular}[c]{@{}l@{}}Die Geschichte der Europäischen Union ist \\ durch ein Geflecht konkurrierender Motive und \\ Entwicklungstendenzen charakterisiert, die zu \\ unterschiedlichen Zeitpunkten jeweils \\ richtungsgebend auf die Entwicklung der \\ Gemeinschaft eingewirkt haben.\end{tabular}} & \begin{tabular}[c]{@{}l@{}}The history of the European Union is characterized \\ by a web of competing motives and development \\ tendencies, each of which has had a directional \\ impact on the development of the community at \\ different points in time.\end{tabular} \\* \midrule
\textbf{Simple} & \multicolumn{1}{l|}{\begin{tabular}[c]{@{}l@{}}Die Geschichte der Europäischen Union ist durch \\ große Unterschiede von Motiven und \\ Entwicklungen gekennzeichnet. Zu \\ unterschiedlichen Zeitpunkten haben diese \\ Unterschiede auf die Entwicklung der \\ Gesellschaft Einfluss gehabt.\end{tabular}} & \begin{tabular}[c]{@{}l@{}}The history of the European Union is marked by \\ great differences in motives and developments. \\ At different points in time, these differences have \\ had an impact on the development of society.\end{tabular} \\* \midrule
\multicolumn{3}{c}{\textbf{WikiAuto (English)}} \\* \midrule
\textbf{Original} & \multicolumn{1}{l|}{\begin{tabular}[c]{@{}l@{}}The news of Kalākaua's death did not reach Hawaii \\ until January 29 when the "Charleston" \\ returned to Honolulu with the king's remains.\end{tabular}} &  \\* \midrule
\textbf{Simple} & \multicolumn{1}{l|}{\begin{tabular}[c]{@{}l@{}}Kalākaua's remains were sent to Honolulu aboard \\ the American cruiser USS "Charleston".\end{tabular}} & \multicolumn{1}{c}{} \\* \midrule
\multicolumn{3}{c}{\textbf{WikiLargeFR (French)}} \\* \midrule
\textbf{Original} & \multicolumn{1}{l|}{\begin{tabular}[c]{@{}l@{}}La couleur du corps varie du brun moyen au doré \\ à blanc beige et, à l'occasion, elle est marquée \\ de taches brun foncé, surtout sur les membres.\end{tabular}} & \begin{tabular}[c]{@{}l@{}}Body color ranges from medium brown to golden \\ to tan-white, and occasionally marked with dark \\ brown spots, especially on the limbs.\end{tabular} \\* \midrule
\textbf{Simple} & \multicolumn{1}{l|}{\begin{tabular}[c]{@{}l@{}}La couleur du corps varie de brun moyen à doré \\ à blanc beige et parfois marquée de taches brun \\ foncé.\end{tabular}} & \begin{tabular}[c]{@{}l@{}}Body color ranges from medium brown to golden \\ to tan-white and sometimes marked with dark \\ brown spots.\end{tabular} \\* \bottomrule
\caption{Example sentences sampled from all of the datasets in the {\sc MultiSim} benchmark}
\label{table:dataset-examples}
\end{longtable}
}
\twocolumn
\clearpage

\end{document}